\documentclass{article}

\usepackage[final]{corl_2018}

\usepackage{amssymb,amsmath,array}
\usepackage{url,balance}
\usepackage{times}
\usepackage{graphicx}
\usepackage{tabularx}
\usepackage{booktabs}
\usepackage{xcolor}
\usepackage{changebar}
\usepackage{subcaption}
\usepackage{natbib}
\title{Continual Lifelong Learning with Neural Networks: A Review}

\author{German I. Parisi$^1$, Ronald Kemker$^2$, Jose L. Part$^3$, Christopher Kanan$^2$, Stefan Wermter$^1$\\
  $^1$Knowledge Technology, Department of Informatics, Universit\"at Hamburg, Germany\\
  $^2$Chester F. Carlson Center for Imaging Science, Rochester Institute of Technology, NY, USA\\
  $^3$Department of Computer Science, Heriot-Watt University, Edinburgh Centre for Robotics, Scotland, UK\\
}

\begin{document}
\maketitle

%===============================================================================

\begin{abstract}
Humans and animals have the ability to continually acquire, fine-tune, and transfer knowledge and skills throughout their lifespan.
This ability, referred to as lifelong learning, is mediated by a rich set of neurocognitive mechanisms that together contribute to the development and specialization of our sensorimotor skills as well as to long-term memory consolidation and retrieval.
Consequently, lifelong learning capabilities are crucial for computational systems and autonomous agents interacting in the real world and processing continuous streams of information.
However, lifelong learning remains a long-standing challenge for machine learning and neural network models since the continual acquisition of incrementally available information from non-stationary data distributions generally leads to catastrophic forgetting or interference.
This limitation represents a major drawback for state-of-the-art deep neural network models that typically learn representations from stationary batches of training data, thus without accounting for situations in which information becomes incrementally available over time.
In this review, we critically summarize the main challenges linked to lifelong learning for artificial learning systems and compare existing neural network approaches that alleviate, to different extents, catastrophic forgetting.
Although significant advances have been made in domain-specific learning with neural networks, extensive research efforts are required for the development of robust lifelong learning on autonomous agents and robots.
We discuss well-established and emerging research motivated by lifelong learning factors in biological systems such as structural plasticity, memory replay, curriculum and transfer learning, intrinsic motivation, and multisensory integration.
\end{abstract}

% Two or three meaningful keywords should be added here

%===============================================================================

\section{Introduction}
	
Computational systems operating in the real world are exposed to continuous streams of information and thus are required to learn and remember multiple tasks from dynamic data distributions.
For instance, an autonomous agent interacting with the environment is required to learn from its own experiences and must be capable of progressively acquiring, fine-tuning, and transferring knowledge over long time spans.
The ability to continually learn over time by accommodating new knowledge while retaining previously learned experiences is referred to as \textit{continual} or \textit{lifelong learning}.
Such a continuous learning task has represented a long-standing challenge for machine learning and neural networks and, consequently, for the development of artificial intelligence (AI) systems~\citep{Hassabis2017, Thrun1995}.

The main issue of computational models regarding lifelong learning is that they are prone to \textit{catastrophic forgetting} or \textit{catastrophic interference}, i.e., training a model with new information interferes with previously learned knowledge~\citep{McClelland1995, McCloskey1989}.
This phenomenon typically leads to an abrupt performance decrease or, in the worst case, to the old knowledge being completely overwritten by the new one.
Current deep neural network learning models excel at a number of classification tasks by relying on a large batch of (partially) annotated training samples (see \cite{Guo2016, LeCun2015} for reviews).
However, such a learning scheme assumes that all samples are available during the training phase and, therefore, requires the retraining of the network parameters on the entire dataset in order to adapt to changes in the data distribution.
When trained on sequential tasks, the performance of conventional neural network models significantly decreases on previously learned tasks as new tasks are learned~\citep{Kemker2018a, Lomonaco2018}.
Although retraining from scratch pragmatically addresses catastrophic forgetting, this methodology is very inefficient and hinders the learning of novel data in real time.
For instance, in scenarios of developmental learning where autonomous agents learn by actively interacting with the environment, there may be no distinction between training and test phases, requiring the learning model to concurrently adapt and timely trigger behavioural responses~\citep{Cangelosi2015, Tani2016}.

For overcoming catastrophic forgetting, learning systems must, on the one hand, show the ability to acquire new knowledge and refine existing knowledge on the basis of the continuous input and, on the other hand, prevent the novel input from significantly interfering with existing knowledge.
The extent to which a system must be plastic in order to integrate novel information and stable in order not to catastrophically interfere with consolidated knowledge is known as the \textit{stability-plasticity dilemma} and has been widely studied in both biological systems and computational models~\citep{Ditzler2015, Mermillod2013, Grossberg1980, Grossberg2012}.
Due to the very challenging but high-impact aspects of lifelong learning, a large body of computational approaches have been proposed that take inspiration from the biological factors of learning from the mammalian brain.

Humans and other animals excel at learning in a lifelong manner, making the appropriate decisions on the basis of sensorimotor contingencies learned throughout their lifespan~\citep{Tani2016, Bremner2012}.
The ability to incrementally acquire, refine, and transfer knowledge over sustained periods of time is mediated by a rich set of neurophysiological processing principles that together contribute to the early development and experience-driven specialization of perceptual and motor skills~\citep{Zenke2017a, Power2016, Murray2016, Lewkowicz2014}.
In Section 2, we introduce a set of widely studied biological aspects of lifelong learning and their implications for the modelling of biologically motivated neural network architectures.
First, we focus on the mechanisms of neurosynaptic plasticity that regulate the stability-plasticity balance in multiple brain areas~(Sec. 2.2 and 2.3).
Plasticity is an essential feature of the brain for neural malleability at the level of cells and circuits (see \cite{Power2016} a survey).
For a stable continuous lifelong process, two types of plasticity are required: (i) Hebbian plasticity~\citep{Hebb1949} for positive feedback instability, and (ii) compensatory homeostatic plasticity which stabilizes neural activity.
It has been observed experimentally that specialized mechanisms protect knowledge about previously learned tasks from interference encountered during the learning of novel tasks by decreasing rates of synaptic plasticity~\citep{Cichon2015}.
Together, Hebbian learning and homeostatic plasticity stabilize neural circuits to shape optimal patterns of experience-driven connectivity, integration, and functionality~\citep{Zenke2017a, Abraham2005}.

Importantly, the brain must carry out two complementary tasks: generalize across experiences and retain specific memories of episodic-like events.
In Section 2.4, we summarize the complementary learning systems (CLS) theory~\citep{McClelland1995, Kumaran2016} which holds the means for effectively extracting the statistical structure of perceived events (generalization) while retaining episodic memories, i.e., the collection of experiences at a particular time and place.
The CLS theory defines the complementary contribution of the hippocampus and the neocortex in learning and memory, suggesting that there are specialized mechanisms in the human cognitive system for protecting consolidated knowledge.
The hippocampal system exhibits short-term adaptation and allows for the rapid learning of new information which will, in turn, be transferred and integrated into the neocortical system for its long-term storage.
The neocortex is characterized by a slow learning rate and is responsible for learning generalities.
However, additional studies in learning tasks with human subjects~\citep{Mareschal2007, Pallier2003} observed that, under certain circumstances, catastrophic forgetting may still occur (see Sec. 2.4).

Studies on the neurophysiological aspects of lifelong learning have inspired a wide range of machine learning and neural network approaches.
In Section 3, we introduce and compare computational approaches that address catastrophic forgetting.
We focus on recent learning models that i) regulate intrinsic levels of synaptic plasticity to protect consolidated knowledge (Sec. 3.2); ii) allocate additional neural resources to learn new information (Sec. 3.3), and iii) use complementary learning systems for memory consolidation and experience replay (Sec. 3.4).
The vast majority of these approaches are designed to address lifelong supervised learning on annotated datasets of finite size (e.g., \cite{Zenke2017b, Kirkpatrick2017} and do not naturally extend to more complex scenarios such as the processing of partially unlabelled sequences.
Unsupervised lifelong learning, on the other hand, has been proposed mostly through the use of self-organizing neural networks (e.g., \cite{Parisi2018b, Parisi2017a, Richardson2008}).
Although significant advances have been made in the design of learning methods with structural regularization or dynamic architectural update, considerably less attention has been given to the rigorous evaluation of these algorithms in lifelong and incremental learning tasks.
Therefore, in Sec. 3.5 we discuss the importance of using and designing quantitative metrics to measure catastrophic forgetting with large-scale datasets.

Lifelong learning has recently received increasing attention due to its implications in autonomous learning agents and robots.
Neural network approaches are typically designed to incrementally adapt to modality-specific, often synthetic, data samples collected in controlled environments, shown in isolation and random order.
This differs significantly from the more ecological conditions humans and other animals are exposed to throughout their lifespan~\citep{Cangelosi2015, Krueger2009, Wermter2005, Skinner1958}.
Agents operating in the real world must deal with sensory uncertainty, efficiently process continuous streams of multisensory information, and effectively learn multiple tasks without catastrophically interfering with previously learned knowledge.
Intuitively, there is a huge gap between the above-mentioned neural network models and more sophisticated lifelong learning agents expected to incrementally learn from their continuous sensorimotor experiences.

Humans can easily acquire new skills and transfer knowledge across domains and tasks~\citep{Barnett2002} while artificial systems are still in their infancy regarding what is referred to as \textit{transfer learning}~\citep{Weiss2016}.
Furthermore, and in contrast with the predominant tendency to train neural network approaches with uni-sensory (e.g., visual or auditory) information, the brain benefits significantly from the integration of multisensory information, providing the means for an efficient interaction also in situations of sensory uncertainty~\citep{Stein2014, Bremner2012, Spence2010}.
The multisensory aspects of early development and sensorimotor specialization in the brain have inspired a large body of research on autonomous embodied agents~\citep{Lewkowicz2014, Cangelosi2015}.
In Section 4, we review computational approaches motivated by biological aspects of learning which include critical developmental stages and curriculum learning~(Sec. 4.2), transfer learning for the reuse of knowledge during the learning of new tasks~(Sec. 4.3), reinforcement learning for the autonomous exploration of the environment driven by intrinsic motivation and self-supervision~(Sec. 4.4), and multisensory systems for crossmodal lifelong learning (Sec. 4.5).

This review complements previous surveys on catastrophic forgetting in connectionist models~\citep{French1999, Goodfellow2013, Soltoggio2017} that do not critically compare recent experimental work (e.g., deep learning) or define clear guidelines on how to train and evaluate lifelong approaches on the basis of experimentally observed developmental mechanisms.
Together, our and previous reviews highlight lifelong learning as a highly interdisciplinary challenge.
Although the individual disciplines may have more open questions than answers, the combination of these findings may provide a breakthrough with respect to current ad-hoc approaches, with neural networks being the stepping stone towards the increasingly sophisticated cognitive abilities exhibited by AI systems.
In Section 5, we summarize the key ideas presented in this review and provide a set of ongoing and future research directions.

\section{Biological Aspects of Lifelong Learning}

\subsection{The Stability-Plasticity Dilemma}

As humans, we have an astonishing ability to adapt by effectively acquiring knowledge and skills, refining them on the basis of novel experiences, and transferring them across multiple domains~\citep{Bremner2012, Calvert2004, Barnett2002}.
While it is true that we tend to gradually forget previously learned information throughout our lifespan, only rarely does the learning of novel information catastrophically interfere with consolidated knowledge~\citep{French1999}.
For instance, the human somatosensory cortex can assimilate new information during motor learning tasks without disrupting the stability of previously acquired motor skills~\citep{Braun2001}.
Lifelong learning in the brain is mediated by a rich set of neurophysiological principles that regulate the stability-plasticity balance of the different brain areas and that contribute to the development and specialization of our cognitive system on the basis of our sensorimotor experiences~\citep{Zenke2017a, Power2016, Murray2016, Lewkowicz2014}.
The stability-plasticity dilemma regards the extent to which a system must be prone to integrate and adapt to new knowledge and, importantly, how this adaptation process should be compensated by internal mechanisms that stabilize and modulate neural activity to prevent catastrophic forgetting~\citep{Ditzler2015, Mermillod2013}

Neurosynaptic plasticity is an essential feature of the brain yielding physical changes in the neural structure and allowing us to learn, remember, and adapt to dynamic environments~(see \cite{Power2016} for a survey).
The brain is particularly plastic during critical periods of early development in which neural networks acquire their overarching structure driven by sensorimotor experiences.
Plasticity becomes less prominent as the biological system stabilizes through a well-specified set of developmental stages, preserving a certain degree of plasticity for its adaptation and reorganization at smaller scales~\citep{Hensch1998, Quadrato2014, Kiyota2017}.
The specific profiles of plasticity during critical and post-developmental periods vary across biological systems~\citep{Uylings2006}, showing a consistent tendency to decreasing levels of plasticity with increasing age~\citep{Hensch2004}.
Plasticity plays a crucial role in the emergence of sensorimotor behaviour by complementing genetic information which provides a specific evolutionary path~\citep{Grossberg2012}.
Genes or molecular gradients drive the initial development for granting a rudimentary level of performance from the start whereas extrinsic factors such as sensory experience complete this process for achieving higher structural complexity and performance~\citep{Hirsch1970, Shatz1996, Sur2001}.
In this review, we focus on the developmental and learning aspects of brain organization while we refer the reader to \cite{Soltoggio2017} for a review of evolutionary imprinting.

\subsection{Hebbian Plasticity and Stability}

The ability of the brain to adapt to changes in its environment provides vital insight into how connectivity and function of the cortex are shaped.
It has been shown that while rudimentary patterns of connectivity in the visual system are established in early development, normal visual input is required for the correct development of the visual cortex.
The seminal work of \cite{Hubel1967} on the emergence of ocular dominance showed the importance of timing of experience on the development of normal patterns of cortical organization.
The visual experience of newborn kittens was experimentally manipulated to study the effects of varied input on brain organization.
The disruption of cortical organization was more severe when the deprivation of visual input began prior to ten weeks of age while no changes were observed in adult animals.
Additional experiments showed that neural patterns of cortical organization can be driven by external environmental factors at least for a period early in development~\citep{Hubel1962, Hubel1970, Hubel1977}.

\begin{figure*}[t]
\centering
\includegraphics[width=\textwidth]{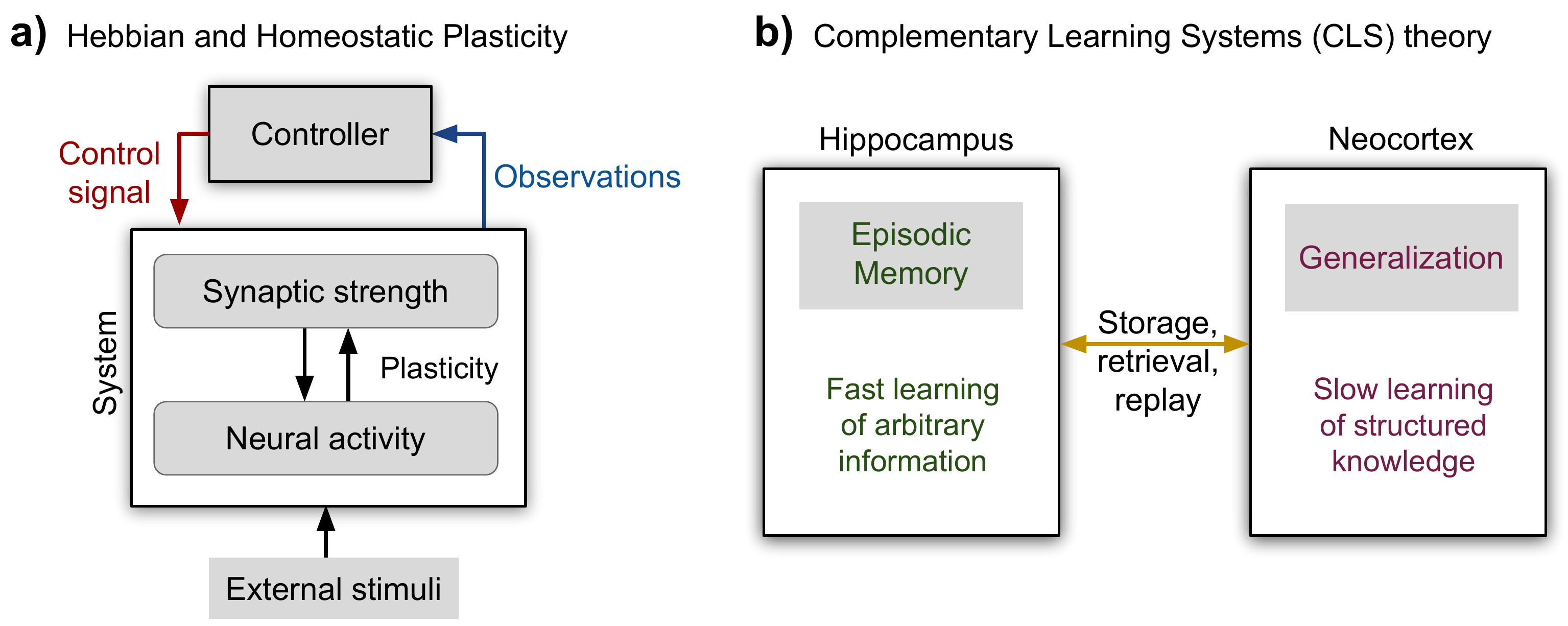}
\caption{Schematic view of two aspects of neurosynaptic adaptation: a) Hebbian learning with homeostatic plasticity as a compensatory mechanism that uses observations to compute a feedback control signal (Adapted with permission from ~\cite{Zenke2017a}). b) The complementary learning systems (CLS) theory~\citep{McClelland1995} comprising the hippocampus for the fast learning of episodic information and the neocortex for the slow learning of structured knowledge.}% (\emph{$G^C$}).}
\label{fig:learning}
\end{figure*}

The most well-known theory describing the mechanisms of synaptic plasticity for the adaptation of neurons to external stimuli was first proposed by~\cite{Hebb1949}, postulating that when one neuron drives the activity of another neuron, the connection between them is strengthened.
More specifically, the Hebb's rule states that the repeated and persistent stimulation of the postsynaptic cell from the presynaptic cell leads to an increased synaptic efficacy.
Throughout the process of development, neural systems stabilize to shape optimal functional patterns of neural connectivity.
The simplest form of Hebbian plasticity considers a synaptic strength $w$ which is updated by the product of a pre-synaptic activity $x$ and the post-synaptic activity $y$:
\begin{equation}
\label{hebbs}
\Delta w = x \cdot y \cdot \eta,
\end{equation}
where $\eta$ is a given learning rate.
However, Hebbian plasticity alone is unstable and leads to runaway neural activity, thus requiring compensatory mechanisms to stabilize the learning process~\citep{Abbott2000, Bienenstock1982}.
Stability in Hebbian systems is typically achieved by augmenting Hebbian plasticity with additional constraints such as upper limits on the individual synaptic weights or average neural activity~\citep{Miller1994, Song2000}.
Homeostatic mechanisms of plasticity include synaptic scaling and meta-plasticity which directly affect synaptic strengths~\citep{Davis2006, Turrigiano2011}.
Without loss of generality, homeostatic plasticity can be viewed as a modulatory effect or feedback control signal that regulates the unstable dynamics of Hebbian plasticity (see Fig.~\ref{fig:learning}.a).
The feedback controller directly affects synaptic strength on the basis of the observed neural activity and must be fast in relation to the timescale of the unstable system~\citep{Astrom2010}.
In its simplest form, modulated Hebbian plasticity can be modelled by introducing an additional modulatory signal $m$ to Eq.~\ref{hebbs} such that the synaptic update is given by
\begin{equation}
\label{hebbs_mod}
\Delta w = m \cdot x \cdot y \cdot \eta.
\end{equation}
Modulatory feedback in Hebbian neural networks has received increasing attention, with different approaches proposing biologically plausible learning through modulatory loops~\citep{Grant2017, Soltoggio2017}.
For a critical review of the temporal aspects of Hebbian and homeostatic plasticity, we refer the reader to \cite{Zenke2017a}.

Evidence on cortical function has shown that neural activity in multiple brain areas results from the combination of bottom-up sensory drive, top-down feedback, and prior knowledge and expectations \citep{Heeger2017}.
In this setting, complex neurodynamic behaviour can emerge from the dense interaction of hierarchically arranged neural circuits in a self-organized manner~\citep{Tani2016}.
Input-driven self-organization plays a crucial role in the brain~\cite{Nelson2000}, with topographic maps being a common feature of the cortex for processing sensory input~\citep{Willshaw1976}.
Different models of neural self-organization have been proposed that resemble the dynamics of basic biological findings on Hebbian-like learning and plasticity~\citep{Kohonen1982, Martinetz1993, Fritzke1995, Marsland2002}, demonstrating that neural map organization results from unsupervised, statistical learning with nonlinear approximations of the input distribution.

To stabilize the unsupervised learning process, neural network self-organization can be complemented with top-down feedback such as task-relevant signals that modulate the intrinsic map plasticity~\citep{Parisi2018b, Soltoggio2017}.
In a hierarchical processing regime, neural detectors have increasingly large spatio-temporal receptive fields to encode information over larger spatial and temporal scales~\citep{Taylor2015, Hasson2008}.
Thus, higher-level layers can provide the top-down context for modulating the bottom-up sensory drive in lower-level layers.
For instance, bottom-up processing is responsible for encoding the co-occurrence statistics of the environment while error-driven signals modulate this feedforward process according to top-down, task-specific factors~\citep{Murray2016}.
Together, these models contribute to a better understanding of the underlying neural mechanisms for the development of hierarchical cortical organization.

\subsection{The Complementary Learning Systems}

The brain learns and memorizes.
The former task is characterized by the extraction of the statistical structure of the perceived events with the aim to generalize to novel situations.
The latter, conversely, requires the collection of separated episodic-like events.
Consequently, the brain must comprise a mechanism to concurrently generalize across experiences while retaining episodic memories.

Sophisticated cognitive functions rely on canonical neural circuits replicated across multiple brain areas~\citep{Douglas1995}.
However, although there are shared structural properties, different brain areas operate at multiple timescales and learning rates, thus differing significantly from each other in a functional way~\citep{Benna2016, Fusi2005}.
A prominent example is the complementary contribution of the neocortex and the hippocampus in learning and memory consolidation~\citep{McClelland1995, OReilly2002, OReilly2004}.
The complementary learning systems (CLS) theory~\citep{McClelland1995} holds that the hippocampal system exhibits short-term adaptation and allows for the rapid learning of novel information which will, in turn, be played back over time to the neocortical system for its long-term retention (see Fig.~\ref{fig:learning}.b).
More specifically, the hippocampus employs a rapid learning rate and encodes sparse representations of events to minimize interference.
Conversely, the neocortex is characterized by a slow learning rate and builds overlapping representations of the learned knowledge.
Therefore, the interplay of hippocampal and neocortical functionality is crucial to concurrently learn regularities (statistics of the environment) and specifics (episodic memories).
Both brain areas are known to learn via Hebbian and error-driven mechanisms~\citep{OReilly2000}.
In the neocortex, feedback signals will yield task-relevant representations while, in the case of the hippocampus, error-driven modulation can switch its functionally between pattern discrimination and completion for recalling information~\citep{OReilly2004}.

Studies show that adult neurogenesis contributes to the formation of new memories~\citep{Altman1963, Eriksson1998, Cameron1993, Gage2000}.
It has been debated whether human adults grow significant amounts of new neurons.
Recent research has suggested that hippocampal neurogenesis drops sharply in children to undetectable levels in adulthood~\citep{Sorrells2018}.
On the other hand, other studies suggest that hippocampal neurogenesis sustains human-specific cognitive function throughout life~\citep{Boldrini2018}.
During neurogenesis, the hippocampus' dentate gyrus uses new neural units to quickly assimilate and immediately recall new information~\citep{Altman1963, Eriksson1998}.
During initial memory formation, the new neural progenitor cells exhibit high levels of plasticity; and as time progresses, the plasticity decreases to make the new memory more stable~\citep{Deng2010}.
In addition to neurogenesis, neurophysiological studies evidence the contribution of synaptic rewiring by structural plasticity on memory formation in adults~\citep{Knoblauch2014, Knoblauch2017}, with a major role of structural plasticity in increasing information storage efficiency in terms of space and energy demands.

While the hippocampus is normally associated with the immediate recall of recent memories (i.e., short-term memories), the prefrontal cortex (PFC) is usually associated with the preservation and recall of remote memories (i.e., long-term memories; \cite{Bontempi1999}).
\cite{Kitamura2017} showed that, when the brain learns something new, the hippocampus and PFC are both initially encoded with the corresponding memory; however, the hippocampus is primarily responsible for the recent recall of new information.
Over time, they showed that the corresponding memory is consolidated over to PFC, which will then take over responsibility for recall of the (now) remote memory.
It is believed that the consolidation of recent memories into long-term storage occurs during rapid eye movement (REM) sleep~\citep{Taupin2002, Gais2007}.

Recently, the CLS theory was updated to incorporate additional findings from neuroscience~\citep{Kumaran2016}.
The first set of findings regards the role of the replaying of memories stored in the hippocampus as a mechanism that, in addition to the integration of new information, also supports the goal-oriented manipulation of experience statistics~\citep{ONeill2010}.
The hippocampus rapidly encodes episodic-like events that can be reactivated during sleep or unconscious and conscious memory recall~\citep{GelbardSagiv2008}, thus consolidating information in the neocortex via the reactivation of encoded experiences in terms of multiple internally generated replays~\citep{Ratcliff1990}.
Furthermore, evidence suggests that (i) the hippocampus supports additional forms of generalization through the recurrent interaction of episodic memories~\citep{Kumaran2012} and (ii) if the new information is consistent with existing knowledge, then its integration into the neocortex is faster than originally suggested~\citep{Tse2011}.
Overall, the CLS theory holds the means for effectively generalizing across experiences while retaining specific memories in a lifelong manner.
However, the exact neural mechanisms remain poorly understood.

\subsection{Learning without Forgetting}

The neuroscience findings described in Sec. 2.3 demonstrate the existence of specialized neurocognitive mechanisms for acquiring and protecting knowledge.
Nevertheless, it has been observed that catastrophic forgetting may occur under specific circumstances.
For instance, \cite{Mareschal2007} found an asymmetric interference effect in a sequential category learning task with 3- and 4-month-old infants.
The infants had to learn two categories, \textit{dog} and \textit{cat}, from a series of pictures and would have to later distinguish a novel animal in a subsequent preferential looking task.
Surprisingly, it was observed that infants were able to retain the category \textit{dog} only if it was learned before \textit{cat}.
This asymmetric effect is thought to reflect the relative similarity of the two categories in terms of perceptual structure.

Additional interference effects were observed for long-term knowledge.
\cite{Pallier2003} studied the word recognition abilities of Korean-born adults whose language environment shifted completely from Korean to French after being adopted between the ages of 3 and 8 by French families.
Behavioural tests showed that subjects had no residual knowledge of the previously learned Korean vocabulary.
Functional brain imaging data showed that the response of these subjects while listening to Korean was no different from the response while listening to other foreign languages that they had been exposed to, suggesting that their previous knowledge of Korean was completely overwritten.
Interestingly, brain activations showed that Korean-born subjects produced weaker responses to French with respect to native French speakers.
It was hypothesized that, while the adopted subjects did not show strong responses to transient exposures to the Korean vocabulary, prior knowledge of Korean may have had an impact during the formulation of language skills to facilitate the re-acquisition of the Korean language should the individuals be re-exposed to it in an immersive way.

Humans do not typically exhibit strong events of catastrophic forgetting because the kind of experiences we are exposed to are very often interleaved~\citep{Seidenberg2006}.
Nevertheless, forgetting effects may be observed when new experiences are strongly immersive such as in the case of children drastically shifting from Korean to French.
Together, these findings reveal a well-regulated balance in which, on the one hand, consolidated knowledge must be protected to ensure its long-term durability and avoid catastrophic interference during the learning of novel tasks and skills over long periods of time.
On the other hand, under certain circumstances such as immersive long-term experiences, old knowledge can be overwritten in favour of the acquisition and refinement of new knowledge.

Taken together, the biological aspects of lifelong learning summarized in this section provide insights into how artificial agents could prevent catastrophic forgetting and model graceful forgetting.
In the next sections, we describe and compare an extensive set of neural network models and AI approaches that have taken inspiration from such principles.
In the case of computational systems, however, additional challenges must be faced due to the limitations of learning in restricted scenarios that typically capture very few components of the processing richness of biological systems.

\section{Lifelong Learning and Catastrophic Forgetting in Neural Networks}

\subsection{Lifelong Machine Learning}

Lifelong learning represents a long-standing challenge for machine learning and neural network systems~\citep{Hassabis2017, French1999}.
This is due to the tendency of learning models to catastrophically forget existing knowledge when learning from novel observations~\citep{Thrun1995}.
A lifelong learning system is defined as an adaptive algorithm capable of learning from a continuous stream of information, with such information becoming progressively available over time and where the number of tasks to be learned (e.g., membership classes in a classification task) are not predefined.
Critically, the accommodation of new information should occur without catastrophic forgetting or interference.

In connectionist models, catastrophic forgetting occurs when the new instances to be learned differ significantly from previously observed examples because this causes the new information to overwrite previously learned knowledge in the shared representational resources in the neural network~\citep{French1999, McCloskey1989}.
When learning offline, this loss of knowledge can be recovered because the agent sees the same pseudo-randomly shuffled examples over and over, but this is not possible when the data cannot be shuffled and is observed as a continuous stream.
The effects of catastrophic forgetting have been widely studied for over two decades, especially in networks learned using back-propagation~\citep{Ratcliff1990, Lewandowsky1994} and in the Hopfield networks~\citep{Nadal1986, Burgess1991}.

Early attempts to mitigate catastrophic forgetting typically consisted of memory systems that store previous data and that regularly replay old samples interleaved with samples drawn from the new data~\citep{Robins1993, Robins1995}, and these methods are still used today~\citep{Gepperth2015, Rebuffi2016}.
However, a general drawback of memory-based systems is that they require explicit storage of old information, leading to large working memory requirements.
Furthermore, in the case of a fixed amount of neural resources, specialized mechanisms should be designed that protect consolidated knowledge from being overwritten by the learning of novel information (e.g., \cite{Zenke2017b, Kirkpatrick2017}).
Intuitively, catastrophic forgetting can be strongly alleviated by allocating additional neural resources whenever they are required (e.g., \cite{Parisi2018b, Parisi2017a, Rusu2016, Hertz1991}).
This approach, however, may lead to scalability issues with significantly increased computational efforts for neural architectures that become very large.
Conversely, since in a lifelong learning scenario the number of tasks and samples per task cannot be known a priori, it is non-trivial to predefine a sufficient amount of neural resources that will prevent catastrophic forgetting without strong assumptions on the distribution of the input.
In this setting, three key aspects have been identified for avoiding catastrophic forgetting in connectionist models~\citep{Richardson2008}: (i)~allocating additional neural resources for new knowledge; (ii)~using non-overlapping representations if resources are fixed; and (iii)~interleaving the old knowledge as the new information is represented.

The brain has evolved mechanisms of neurosynaptic plasticity and complex neurocognitive functions that process continuous streams of information in response to both short- and long-term changes in the environment~\citep{Zenke2017a, Power2016, Murray2016, Lewkowicz2014}.
Consequently, the differences between biological and artificial systems go beyond architectural differences, and also include the way in which these artificial systems are exposed to external stimuli.
Since birth, humans are immersed in a highly dynamic world and, in response to this rich perceptual experience, our neurocognitive functions progressively develop to make sense of increasingly more complex events.
Infants start with relatively limited capabilities for processing low-level features and incrementally develop towards the learning of higher-level perceptual, cognitive, and behavioural functions.

Humans make massive use of the spatio-temporal relations and increasingly richer high-order associations of the sensory input to learn and trigger meaningful behavioural responses.
Conversely, artificial systems are typically trained in batches, exposing the learning algorithm to multiple iterations of the same training samples in a (pseudo-)random order.
After a fixed number of training epochs, it is expected that the learning algorithm has tuned its internal representations and can predict novel samples that follow a similar distribution with respect to the training dataset.
Clearly, this approach can be effective (and this is supported by the state-of-the-art performance of deep learning architectures for visual classification tasks; see \cite{Guo2016, LeCun2015} for reviews), but it does not reflect the characteristics of lifelong learning tasks.

\begin{figure*}[t]
\centering
\includegraphics[width=\textwidth]{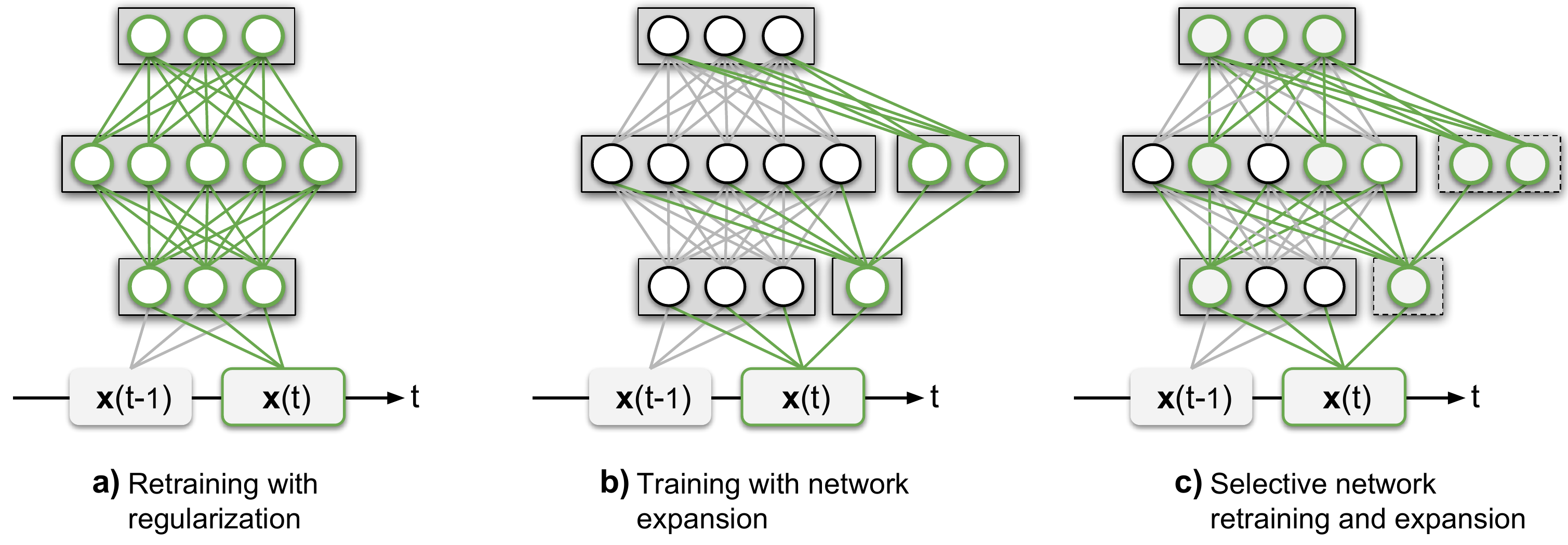}
\caption{Schematic view of neural network approaches for lifelong learning: a) retraining while regularizing to prevent catastrophic forgetting with previously learned tasks, b) unchanged parameters with network extension for representing new tasks, and c) selective retraining with possible expansion.}% (\emph{$G^C$}).}
\label{fig:architectures}
\end{figure*}

In the next sections, we introduce and compare different neural network approaches for lifelong learning that mitigate, to different extents, catastrophic forgetting.
Conceptually, these approaches can be divided into methods that retrain the whole network while regularizing to prevent catastrophic forgetting with previously learned tasks (Fig.~\ref{fig:architectures}.a; Sec. 3.2), methods that selectively train the network and expand it if necessary to represent new tasks (Fig.~\ref{fig:architectures}.b,c; Sec. 3.3), and methods that model complementary learning systems for memory consolidation, e.g. by using memory replay to consolidate internal representations~(Sec. 3.4).
Since considerably less attention has been given to the rigorous evaluation of these algorithms in lifelong learning tasks, in Sec. 3.5 we highlight the importance of using and designing new metrics to measure catastrophic forgetting with large-scale datasets.

\subsection{Regularization Approaches}

Regularization approaches alleviate catastrophic forgetting by imposing constraints on the update of the neural weights.
Such approaches are typically inspired by theoretical neuroscience models suggesting that consolidated knowledge can be protected from forgetting through synapses with a cascade of states yielding different levels of plasticity~\citep{Benna2016, Fusi2005}.
From a computational perspective, this is generally modelled via additional regularization terms that penalize changes in the mapping function of a neural network.

\cite{Li2016} proposed the learning without forgetting (LwF) approach composed of convolutional neural networks (CNN) in which the network with predictions of the previously learned tasks is enforced to be similar to the network with the current task by using knowledge distillation, i.e., the transferring of knowledge from a large, highly regularized model into a smaller model~\citep{Hinton2014}.
According to the LwF algorithm, given a set of shared parameters $\theta_s$ across all tasks, it optimizes the parameters of the new task $\theta_n$ together with $\theta_s$ imposing the additional constraint that the predictions on the samples of the novel task using $\theta_s$ and the parameters of old tasks $\theta_o$ do not shift significantly in order to remember $\theta_o$.
Given the training data on the new task $(X_n,Y_n)$, the output of old tasks for the new data $Y_o$, and randomly initialized new parameters $\theta_n$, the updated parameters $\theta^*_s,\theta^*_o,\theta^*_n$ are given by:
\begin{equation}
\theta^*_s,\theta^*_o,\theta^*_n \leftarrow \text{argmin}_{\hat{\theta}_s,\hat{\theta}_o,\hat{\theta}_n} \left( \lambda_o \mathcal{L}_{old}(Y_o,\hat{Y}_o) + \mathcal{L}_{new}(Y_n,\hat{Y}_n) + \mathcal{R}(\hat{\theta}_s,\hat{\theta}_o,\hat{\theta}_n) \right),
\end{equation}
where $\mathcal{L}_{old}(Y_o,\hat{Y}_o)$ and $\mathcal{L}_{new}(Y_n,\hat{Y}_n)$ minimize the difference between the predicted values $\hat{Y}$ and the ground-truth values $Y$ of the new and old tasks respectively using $\hat{\theta}_s,\hat{\theta}_o,\hat{\theta}_n$, $\lambda_o$ is used to balance new/old tasks, and $\mathcal{R}$ is a regularization term to prevent overfitting.
However, this approach has the drawbacks of highly depending on the relevance of the tasks and that the training time for one task linearly increases with the number of learned tasks.
Additionally, while distillation provides a potential solution to multi-task learning, it requires a reservoir of persistent data for each learned task.
\cite{Jung2018} proposed to regularize the $l_2$ distance between the final hidden activations, preserving the previously learned input-output mappings by computing additional activations with the parameters of the old tasks. 
These approaches, however, are computationally expensive since they require to compute the old tasks' parameters for each novel data sample.
Other approaches opt to either completely prevent the update of weights trained on old tasks~\citep{Razavian2014} or to reduce the learning rate in order to prevent significant changes in the network parameters while training with new data~\citep{Donahue2014}.

\cite{Kirkpatrick2017} proposed the elastic weight consolidation (EWC) model in supervised and reinforcement learning scenarios.
The approach consists of a quadratic penalty on the difference between the parameters for the old and the new tasks that slows down the learning for task-relevant weights coding for previously learned knowledge.
The relevance of the parameter $\theta$ with respect to a task's training data $\mathcal{D}$ is modelled as the posterior distribution $p(\theta \mid \mathcal{D})$.
Assuming a scenario with two independent tasks $A$ with $\mathcal{D}_A$ and $B$ with $\mathcal{D}_B$, the log value of the posterior probability given by the Bayes' rule is:
\begin{equation}
\text{log} p(\theta \mid \mathcal{D}) = \text{log} p(\mathcal{D}_B \mid \theta) + \text{log} p(\theta \mid \mathcal{D}_A) - \text{log} p(\mathcal{D}_B),
\end{equation}
where the posterior probability $\text{log} p(\theta\mid\mathcal{D}_A)$ embeds all the information about the previous task.
However, since this term is intractable, EWC approximates it as a Gaussian distribution with mean given by the parameters $\theta^*_A$ and a diagonal precision given by the diagonal of the fisher information matrix $F$.
Therefore, the loss function of EWC is given by
\begin{equation}
\mathcal{L}(\theta) = \mathcal{L}_B(\theta) + \sum_i \frac{\lambda}{2} F_i (\theta_i - \theta^*_{A,i})^2,
\end{equation}
where $\mathcal{L}_B$ is the loss of $B$, $\lambda$ sets the relevance of the old tasks with respect to the new one, and $i$ denotes the indexes of the parameters.
Therefore, this approach requires a diagonal weighting over the parameters of the learned tasks which is proportional to the diagonal of the Fisher information metric, with the synaptic importance being computed offline and limiting its computational application to low-dimensional output spaces. Furthermore, additional experiments by \cite{Kemker2018a} have shown that, although EWC outperforms other methods for permutation tasks, it is not capable of learning new categories incrementally.

\cite{Zenke2017b} proposed to alleviate catastrophic forgetting by allowing individual synapses to estimate their importance for solving a learned task.
Similar to \cite{Kirkpatrick2017}, this approach penalizes changes to the most relevant synapses so that new tasks can be learned with minimal forgetting.
To reduce large changes in important parameters $\theta_k$ when learning a new task, the authors use a modified cost function $\mathcal{L}^*_n$ with a surrogate loss which approximates the summed loss functions of all previous tasks $\mathcal{L}^*_o$:
\begin{equation}
\mathcal{L}^*_n = \mathcal{L}_n + c \sum_k \Omega^n_k (\theta^*_k-\theta_k)^2,
\end{equation}
where $c$ is a weighting parameter to balance new and old tasks, $\theta^*_k$ are the parameters at the end of the previous task, and $\Omega^n_k$ is a per-parameter regulation strength.
Similar to EWC by \cite{Kirkpatrick2017}, this approach pulls back the more influential parameters towards a reference weight with good performance on previous tasks.
In this case, however, synaptic relevance is computed in an online fashion over the entire learning trajectory in the parameter space.
The two approaches have shown similar results on the Permuted MNIST benchmark~\citep{LeCun1998}.

\cite{Lomonaco2018} proposed the AR1 model for single-incremental-task scenarios which combines architectural and regularization strategies.
Regularization approaches tend to progressively reduce the magnitude of weight changes batch by batch, with most of the changes occurring in the top layers.
Instead, in AR1 intermediate layers weights are adapted without negative impact in terms of forgetting.
Reported results on CORe50~\citep{Lomonaco2017} and iCIFAR-100~\citep{Krizhevsky2009} show that AR1 allows the training of deep convolutional models with less forgetting, outperforming LwF, EWC, and SI.

Ensemble methods have been proposed to alleviate catastrophic forgetting by training multiple classifiers and combine them to generate a prediction.
Early attempts showed a disadvantage linked to the intense use of storage memory which scales up with the number of sessions~\citep{Polikar2001, Dai2007}, while more recent approaches restrict the size of the models through multiple strategies.
For instance, \cite{Ren2017} proposed to adaptively adjust to the changing data distribution by combining sub-models after a new training phase, learning new tasks without referring to previous training data.
\cite{Coop2013} introduced a multi-layer perceptron (MLP) augmented with a fixed expansion layer (FEL) which embeds a sparsely encoding hidden layer to mitigate the interference of previously learned representations.
Ensembles of FEL networks were used to control levels of plasticity, yielding incremental learning capabilities while requiring minimal storage memory.
\cite{Fernando2017} proposed an ensemble method in which a genetic algorithm is used to find the optimal path through a neural network of fixed size for replication and mutation.
This approach, referred to as PathNet, uses agents embedded in a neural network to discover which parts of the network can be reused for the learning of new tasks while freezing task-relevant paths for avoiding catastrophic forgetting.
PathNet's authors showed that incrementally learning new tasks sped up the training of subsequently learned supervised and reinforcement learning tasks; however, they did not measure performance on the original task to determine if catastrophic forgetting occurred.
In addition, PathNet requires an independent output layer for each new task, which prevents it from learning new classes incrementally~\citep{Kemker2018a}.

%He and Jaeger (2018) introduced conceptor-aided backpropagation (CAB), which uses conceptors to shield gradients from forgetting prior knowledge.
%Thus, learning a new task should interfere only minimally with previously learned tasks.
%Furthermore, conceptors provide an analytical tool to discuss the relatedness and enabling conditions for continual learning in terms of the opportunities to re-use previously acquired functionality in subsequent learning episodes.
%Serr{\`a} et al., (2018) incorporated a hard attention (HAT) mask into the loss function and showed that this alleviated catastrophic forgetting in sequential task learning.
%The HAT adds a minimum number of weights to the base network and is trained together with the global model, with a minimal overhead using vanilla stochastic gradient descent.
%Both CAB and HAT outperform EWC on a set of continual learning tasks using the MNIST dataset.

In summary, regularization approaches provide a way to alleviate catastrophic forgetting under certain conditions.
However, they comprise additional loss terms for protecting consolidated knowledge which, with a limited amount of neural resources, may lead to a trade-off on the performance of old and novel tasks.

\subsection{Dynamic Architectures}

The approaches introduced here change architectural properties in response to new information by dynamically accommodating novel neural resources, e.g., re-training with an increased number of neurons or network layers.

For instance, \cite{Rusu2016} proposed to block any changes to the network trained on previous knowledge and expand the architecture by allocating novel sub-networks with fixed capacity to be trained with the new information.
This approach, referred to as \textit{progressive networks}, retains a pool of pre-trained models (one for each learned task $\mathcal{T}_n$).
Given $N$ existing tasks, when a new task is $\mathcal{T}_{N+1}$ is given, a new neural network is created and the lateral connections with the existing tasks are learned.
To avoid catastrophic forgetting, the learned parameters $\theta^n$ for existing tasks $\mathcal{T}_n$ are left unchanged while the new parameter set $\theta^{N+1}$ is learned for $\mathcal{T}_{N+1}$.
Experiments reported good results on a wide variety of reinforcement learning tasks, outperforming common baseline approaches that either pre-train or incrementally fine-tune the models by incorporating prior knowledge only at initialization.
Intuitively, this approach prevents catastrophic forgetting but leads the complexity of the architecture to grow with the number of learned tasks.

\cite{Zhou2012} proposed the incremental training of a denoising autoencoder that adds neurons for samples with high loss and subsequently merges these neurons with existing ones to prevent redundancy.
More specifically, the algorithm is composed of two processes for (i) adding new features to minimize the residual of the objective function and (ii) merging similar features to obtain a compact feature representation and in this way prevent overfitting.
This model was shown to outperform non-incremental denoising autoencoders in classification tasks with the MNIST~\citep{LeCun1998} and the CIFAR-10~\citep{Krizhevsky2009} datasets.
\cite{Cortes2016} proposed to adapt both the structure of the network and its weights by balancing the model complexity and empirical risk minimization.
In contrast to enforcing a pre-defined architecture, the algorithm learns the required model complexity in an adaptive fashion.
The authors reported good results on several binary classification tasks extracted from the CIFAR-10 dataset.
In contrast to previously introduced approaches that do not consider multi-task scenarios, \cite{Xiao2014} proposed a training algorithm with a network that incrementally grows in capacity and also in hierarchical fashion.
Classes are grouped according to their similarity and self-organized into multiple levels, with models inheriting features from existing ones to speed up the learning.
In this case, however, only the topmost layers can grow and the vanilla back-propagation training procedure is inefficient.

%Lopez-Paz and Ranzato (2017) proposed the Gradient Episodic Memory (GEM) model that alleviates catastrophic forgetting and transfers beneficial knowledge to past tasks.
%The model learns the subset of correlations common to a set of distributions or tasks and able to predict target values associated with previous or novel tasks without making use of task descriptors.
%The GEM model does not leverage structure task descriptors which can be used to forward transfer learning.
%Furthermore, each iteration of the GEM algorithm requires one backward pass per task, leading to an increasing computational time as new tasks are learned.
%Experiments with the MNIST and CIFAR100 datasets demonstrate better performance with respect to EWC.

\cite{Draelos2017} incrementally trained an autoencoder on new MNIST digits using the reconstruction error to show whether the older digits were retained.
Their neurogenesis deep learning (NDL) model adds new neural units to the autoencoder to facilitate the addition of new MNIST digits, and it uses intrinsic replay (a generative model used for pseudo-rehearsal) to preserve the weights required to retain older information.
%Lee et al. (2018) expanded this concept to the supervised learning paradigm by creating a dynamically expanding network (i.e., create new units) that uses selective retraining (i.e., only re-train weights affected by a new class) for incremental class learning.
\cite{Yoon2018} took this concept to the supervised learning paradigm and proposed a dynamically expanding network (DEN) that increases the number of trainable parameters to incrementally learn new tasks.
DEN is trained in an online manner by performing selective retraining which expands the network capacity
using group sparse regularization to decide how many neurons to add at each layer.
%Their framework yielded superior performance, compared to EWC, on the experiments outlined in Kirkpatrick et al. (2017).

\cite{Part2016, Part2017} proposed the combination of a pre-trained CNN with a self-organizing incremental neural network (SOINN) in order to take advantage of the good representational power of CNNs and, at the same time, allow the classification network to grow according to the task requirements in a continuous object recognition scenario.
An issue that arises from these types of approaches is scalability since the classification network grows with the number of classes that have been learned.
Another problem that was identified through this approach is that by relying on fixed representations, e.g., pre-trained CNNs, the discrimination power will be conditioned by the dataset used to train the feature extractor.
\cite{Rebuffi2016} deal with this problem by storing example data points that are used along with new data to dynamically adapt the weights of the feature extractor, a technique that is referred to as rehearsal.
By combining new and old data, they prevent catastrophic forgetting but at the expense of a higher memory footprint.

So far, we have considered approaches designed for (or at least strictly evaluated on) the classification of static images.
However, in more natural learning scenarios, sequential input underlying spatio-temporal relations such as in the case of videos must be accounted for.
\cite{Parisi2017a} showed that lifelong learning of human action sequences can be achieved in terms of prediction-driven neural dynamics with internal representations emerging in a hierarchy of recurrent self-organizing networks.
The self-organizing networks can dynamically allocate neural resources and update connectivity patterns according to competitive Hebbian learning.
Each neuron of the neural map consists of a weight vector $\textbf{w}_j$ and a number $K$ of context descriptors $\textbf{c}_{k,j}$ with $\textbf{w}_j,\textbf{c}_{k,j}\in\mathbb{R}^n$.
As a result, recurrent neurons in the map will encode prototype sequence-selective snapshots of the input.
Given a set of recurrent neurons, $N$, the best-matching unit (BMU) $\textbf{w}_b$ with respect to the input $\textbf{x}(t)\in\mathbb{R}^n$ is computed as:
\begin{equation} \label{eq:GetB}
b = \arg\min_{j\in N} \left( \alpha_0 \Vert \textbf{x}(t) - \textbf{w}_j  \Vert^2 + \sum_{k=1}^{K}\ \alpha_k \Vert \textbf{C}_k(t)-\textbf{c}_{j, k}\Vert^2 \right),
\end{equation}
where $\{\alpha_i\}_{i=0...K}$ are constant values that modulate the influence of the current input with respect to previous neural activity and $\textbf{C}_{k}(t)\in\mathbb{R}^n$ is the global context of the network.
Each neuron is equipped with a habituation counter $h_i$ expressing how frequently it has fired based on a simplified model of how the efficacy of a habituating synapse reduces over time.
The network is initialized with two neurons and, at each learning iteration, it inserts a new neuron whenever the activity of the network of a habituated neuron is smaller than a given threshold.
The neural update rule is given by:
\begin{equation}\label{eq:UpdateRateW}
\Delta \textbf{w}_i = \epsilon_i \cdot h_i \cdot (\textbf{x}(t) - \textbf{w}_i),
\end{equation}
where $\epsilon_i$ is a constant learning rate and $h_i$ acts as a modulatory factor (see Eq.~\ref{hebbs_mod}) that decreases the magnitude of learning over time to protect consolidated knowledge.
This approach has shown competitive results with batch learning methods on the Weizmann~\citep{Gorelick2005} and the KTH~\citep{Schuldt2004} action benchmark datasets.
Furthermore, it learns robust action-label mappings also in the case of occasionally missing or corrupted class labels.
\cite{Parisi2018c} showed that self-organizing networks with additive neurogenesis show a better performance than a static network with the same number of neurons, thereby providing insights into the design of neural architectures in incremental learning scenarios when the total number of neurons is fixed.

Similar GWR-based approaches have been proposed for the incremental learning of body motion patterns~\citep{Mici2017, Elfaramawy2017, Parisi2016} and human-object interaction~\citep{Mici2018}.
However, these unsupervised learning approaches do not take into account top-down task-relevant signals that can regulate the stability-plasticity balance, potentially leading to scalability issues for large-scale datasets.
To address this issue, task-relevant modulatory signals were modelled by \cite{Parisi2018b} which regulate the process of neurogenesis and neural update (see Sec.~3.4).
This model shares a number of conceptual similarities with the adaptive resonance theory (ART; see \cite{Grossberg2012} for a review) in which neurons are iteratively adapted to a non-stationary input distribution in an unsupervised fashion and new neurons can be created in correspondence of dissimilar input data.
In the ART model, learning occurs through the interaction of top-down and bottom-up processes: top-down expectations act as memory templates (or prototypes) which are compared to bottom-up sensory observations.
Similar to the GWR's activation threshold, the ART model uses a vigilance parameter to produce fine-grained or more general memories.
Despite its inherent ability to mitigate catastrophic forgetting during incremental learning, an extensive evaluation with recent lifelong learning benchmarks has not been reported for continual learning tasks.
However, it has been noted that the results of some variants of the ART model depend significantly upon the order in which the training data are processed.

While the mechanisms for creating new neurons and connections in the GWR do not resemble biologically plausible mechanisms (e.g., \cite{Eriksson1998, Ming2011, Knoblauch2017}), the GWR learning algorithm represents an efficient computational model that incrementally adapts to non-stationary input.
Crucially, the GWR model creates new neurons whenever they are required and only after the training of existing ones.
The neural update rate decreases as the neurons become more habituated, which has the effect of preventing that noisy input interferes with consolidated neural representations.
Alternative theories suggest that an additional function of hippocampal neurogenesis is the encoding of time for the formation of temporal associations in memory~\citep{Aimone2006, Aimone2009}, e.g., in terms of temporal clusters of long-term episodic memories.
Although the underlying mechanisms of neurogenesis and structural plasticity remain to be further investigated in biological systems, these results reinforce that growing neural models with plasticity constitute effective mitigation of catastrophic forgetting in non-stationary environments.

\subsection{Complementary Learning Systems and Memory Replay}

The CLS theory~\citep{McClelland1995, Kumaran2016} provides the basis for a computational framework modelling memory consolidation and retrieval in which the complementary tasks of memorization and generalization are mediated by the interplay of the mammalian hippocampus and neocortex~(see~Sec. 2.3).
Importantly, the interplay of an episodic memory (specific experience) and a semantic memory (general structured knowledge) provides important insights into the mechanisms of knowledge consolidation in the absence of sensory input.

Dual-memory learning systems have taken inspiration, to different extents, from the CLS theory to address catastrophic forgetting.
An early computational example of this concept was proposed by \cite{Hinton1987} in which each synaptic connection has two weights: a plastic weight with slow change rate which stores long-term knowledge and a fast-changing weight for temporary knowledge.
This dual-weight method reflects the properties of complementary learning systems to mitigate catastrophic forgetting during sequential task learning.
\cite{French1997} developed a pseudo-recurrent dual-memory framework, one for early processing and the other for long-term storage, that used pseudo-rehearsal~\citep{Robins1995} to transfer memories between memory centers.
In pseudo-rehearsal, training samples are not explicitly kept in memory but drawn from a probabilistic model.
During the next two decades, numerous neural network approaches based on CLS principles were used to explain and predict results in different learning and memory domains (see \cite{OReilly2002} for a review).
However, there is no empirical evidence that shows that these approaches can scale up to a large number of tasks or current image and video benchmark datasets (see Sec. 3.5).

More recently, \cite{Soltoggio2015} proposed the use of short- and long-term plasticity for consolidating new information on the basis of a cause-effect hypothesis testing when learning with delayed rewards.
In this case, the difference between the short- and long-term plasticity is not related to the duration of the memory but rather to the confidence of consistency of cause-effect relationships.
This meta-plasticity rule, referred to as hypothesis testing plasticity (HTP), shows that such relationships can be extracted from ambiguous information flows, thus towards explaining the learning in more complex environments (see Sec. 4).

\cite{Gepperth2015} proposed two approaches for incremental learning using (i) a modified self-organizing map (SOM) and (ii) a SOM extended with a short-term memory (STM).
We refer to these two approaches as GeppNet and GeppNet+STM respectively.
In the case of the GeppNet, task-relevant feedback from a regression layer is used to select whether learning in the self-organizing hidden layer should occur.
In the GeppNet+STM case, the STM is used to store novel knowledge which is occasionally played back to the GeppNet layer during sleep phases interleaved with training phases.
This latter approach yielded better performance and faster convergence in incremental learning tasks with the MNIST dataset.
However, the STM has a limited capacity, thus learning new knowledge can overwrite old one.
In both cases, the learning process is divided into two phases: one for initialization and the other for actual incremental learning.
Additional experiments showed that this approach performs significantly worse than EWC~\citep{Kirkpatrick2017} on different permutation tasks~(see Sec. 3.5).
Both GeppNet and GeppNet+STM require storing the entire training dataset during training.

Inspired by the generative role of the hippocampus for the replay of previously encoded experiences, \cite{Shin2017} proposed a dual-model architecture consisting of a deep generative model and a task solver.
In this way, training data from previously learned tasks can be sampled in terms of generated pseudo-data and interleaved with information from the new tasks.
Thus, it is not necessary to explicitly revise old training samples for experience replay, reducing the requirements of working memory.
This approach is conceptually similar to previous ones using a pseudo-rehearsal method, i.e., interleaving information of a new task with internally generated samples from previously learned tasks.
\cite{Robins1995} showed that interleaving information of new experiences with internally generated patterns of previous experiences help consolidate existing knowledge without explicitly storing training samples.
Pseudo-rehearsal was also used by \cite{Draelos2017} for the incremental training of an autoencoder, using the output statistics of the encoder to generate input for the decoder during the replay.
However, similar to most of the above-described approaches, the use of pseudo-rehearsal methods was strictly evaluated on two datasets of relatively low complexity, e.g. the MNIST and the Street View House Number (SVHN)~\citep{Netzer2011}.
Consequently, the question arises whether this generative approach can scale up to more complex domains.

\cite{Lueders2016} proposed an \textit{evolvable} Neural Turing Machine (ENTM) that enables agents to store long-term memories by progressively allocating additional external memory components.
The optimal structure for a continually learning network is found from an initially minimal configuration by evolving network’s topology and weights.
The ENTM configurations can perform one-shot learning of new associations and mitigate the effects of catastrophic forgetting during incremental learning tasks.
A set of reported experiments in reinforcement learning tasks showed that the dynamic nature of the ENTM approach will cause the agents to continually expand its memory over time.
This can lead to an unnecessary memory expansion that would slow down the learning process significantly.
A possible solution to address this issue can be the introduction of cost functions for a more efficient memory allocation and use.

\cite{LopezPaz2017} proposed the Gradient Episodic Memory (GEM) model that yields positive transfer of knowledge to previous tasks.
The main feature of GEM to minimize catastrophic forgetting is an episodic memory used to store a subset of the observed examples from a given task.
While minimizing the loss on the current task $t$, GEM treats the losses on the episodic memories of tasks $k<t$ as inequality constraints, avoiding their increase but allowing their decrease.
This method requires considerable more memory than other regularization approaches such as EWC~\citep{Kirkpatrick2017} at training time (with an episodic memory $\mathcal{M}_k$ for each task $k$) but can work much better in the single pass setting.

\cite{Kemker2018b} proposed the FearNet model for incremental class learning that is inspired by studies of recall and consolidation in the mammalian brain during fear conditioning~\citep{Kitamura2017}.
FearNet uses a hippocampal network capable of immediately recalling new examples, a PFC network for long-term memories, and a third neural network inspired by the basolateral amygdala for determining whether the system should use the PFC or hippocampal network for a particular example.
FearNet consolidates information from its hippocampal network to its PFC network during sleep phases.
FearNet's PFC model is a generative neural network that creates pseudo-samples that are then intermixed with recently observed examples stored in its hippocampal network.
\cite{Kamra2018} presented a similar dual-memory framework that also uses a variational autoencoder as a generative model for pseudo-rehearsal.
Their framework generates a short-term memory module for each new task; however, prior to consolidation, predictions are made using an oracle (i.e., they know which module contains the associated memory).

\cite{Parisi2018b} proposed a dual-memory self-organizing architecture for learning spatiotemporal representations from videos in a lifelong fashion.
The complementary memories are modelled as recurrent self-organizing neural networks: the episodic memory quickly adapts to incoming novel sensory observations via competitive Hebbian Learning, whereas the semantic memory progressively learns compact representations by using task-relevant signals to regulate intrinsic levels of structural plasticity.
For the consolidation of knowledge in the absence of sensory input, trajectories of neural reactivations from the episodic memory are periodically replayed to both memories.
Reported experiments show that the described method significantly outperforms previously proposed lifelong learning methods in three different incremental learning tasks with the CORe50 benchmark dataset~(\cite{Lomonaco2017}; see Sec.~3.5).
Since the development of the neural maps is unsupervised, this approach can be used in scenarios where the annotations of training samples are sparse.

\subsection{Benchmarks and Evaluation Metrics}

Despite the large number of proposed methods addressing lifelong learning, there is no established consensus on benchmark datasets and metrics for their proper evaluation.
Typically, a direct comparison of different methods is hindered by the highly heterogeneous and often limited evaluation schemes to assess the overall performance, levels of catastrophic forgetting, and knowledge transfer.

\cite{LopezPaz2017} defined training and evaluation protocols to assess the quality of continual learning models in terms of their accuracy as well as their ability to transfer knowledge between tasks.
The transfer of knowledge can be \textit{forwards} or \textit{backwards}.
The former refers to the influence that learning a task $\mathcal{T}_A$ has on the performance of a future task $\mathcal{T}_B$, whereas the latter refers to the influence of a current task $\mathcal{T}_B$ on a previous task $\mathcal{T}_A$.
The transfer is \textit{positive} when learning about $\mathcal{T}_A$ improves the performance of another task $\mathcal{T}_B$ (forwards or backwards) and \textit{negative} otherwise.
(See Sec.~4.3 for an introduction to learning models addressing transfer learning.)

%In Sec. 3.2-4, we critically summarized numerous approaches that alleviate catastrophic forgetting through different architectural or functional methods.
%However, a direct comparison of these methods is hindered by the highly heterogeneous and often limited experimental schemes used to evaluate them.
%In addition to benchmark experimental procedures, quantitative metrics are required to assess the effects of catastrophic forgetting with the aim to provide a comparative evaluation of different methods.
%In the context of unsupervised self-organizing architectures, it has been proposed to assess the quality of the topological maps in terms of neural activity, discrimination, and organization~\citep{Richardson2008, Oliver2000}.
%Conversely, supervised methods generally compare the overall classification performance.
%\cite{LopezPaz2017} introduced two metrics to measure how much influence previously learned tasks have on the current task (\textit{backwards transfer}) and how much the current task will influence the learning of future tasks (\textit{forward transfer}).
%\cite{Kemker2018a} proposed three metrics to assess the ability of models to retain old knowledge and acquire new information.

\cite{Kemker2018a} suggested a set of guidelines for evaluating lifelong learning approaches and performed complementary experiments that provide a direct quantitative comparison of a number of approaches.
Such guidelines comprise the use of three benchmark experiments: (i) \textit{data permutation}, (ii) \textit{incremental class learning}, and (iii) \textit{multimodal learning}.
The data permutation experiment consists in training a model with a dataset along with a permuted version of the same dataset, which tests the model's ability to incrementally learn new information with similar feature representations. 
It is then expected that the model prevents catastrophic forgetting with the original data during the subsequent learning of randomly permuted data samples.
In the incremental class learning experiment, the model performance reflects its ability to retain previously learned information while incrementally learning one class at a time.
Finally, in the multimodal learning experiment, the same model is sequentially trained with datasets of different modalities, which tests the model's ability to incrementally learn new information with dramatically different feature representations (e.g., first learn an image classification dataset and then learn an audio classification dataset).

\begin{figure*}[t]
\centering
\includegraphics[width=\textwidth]{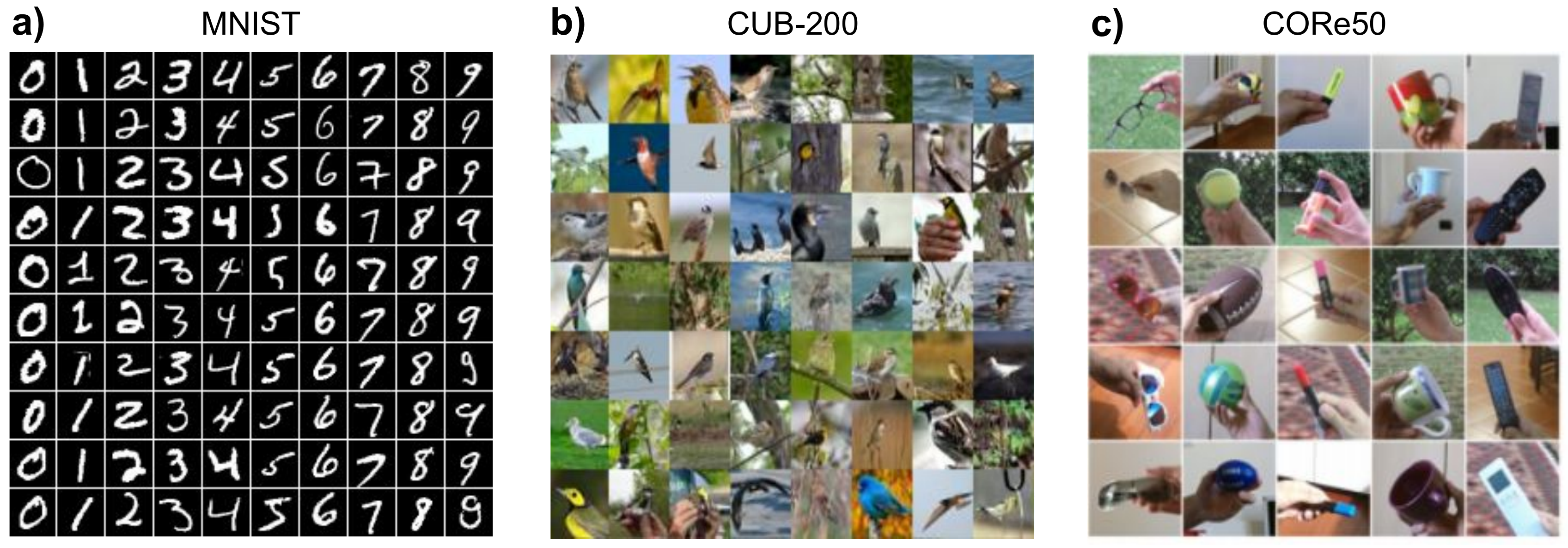}
\caption{Example images from benchmark datasets used for the evaluation of lifelong learning approaches: a) the MNIST dataset with 10 digit classes~\citep{LeCun1998}, b) the Caltech-UCSD Birds-200 (CUB-200) dataset composed of 200 different bird species~\citep{Wah2011}, and c) the CORe50 containing 50 objects with variations in background, illumination, blurring, occlusion, pose, and scale (adapted with permission from \cite{Lomonaco2017}).}% (\emph{$G^C$}).}
\label{fig:datasets}
\end{figure*}

\begin{figure*}
    \centering
    \begin{subfigure}[b]{0.47\textwidth}
        \includegraphics[width=\textwidth]{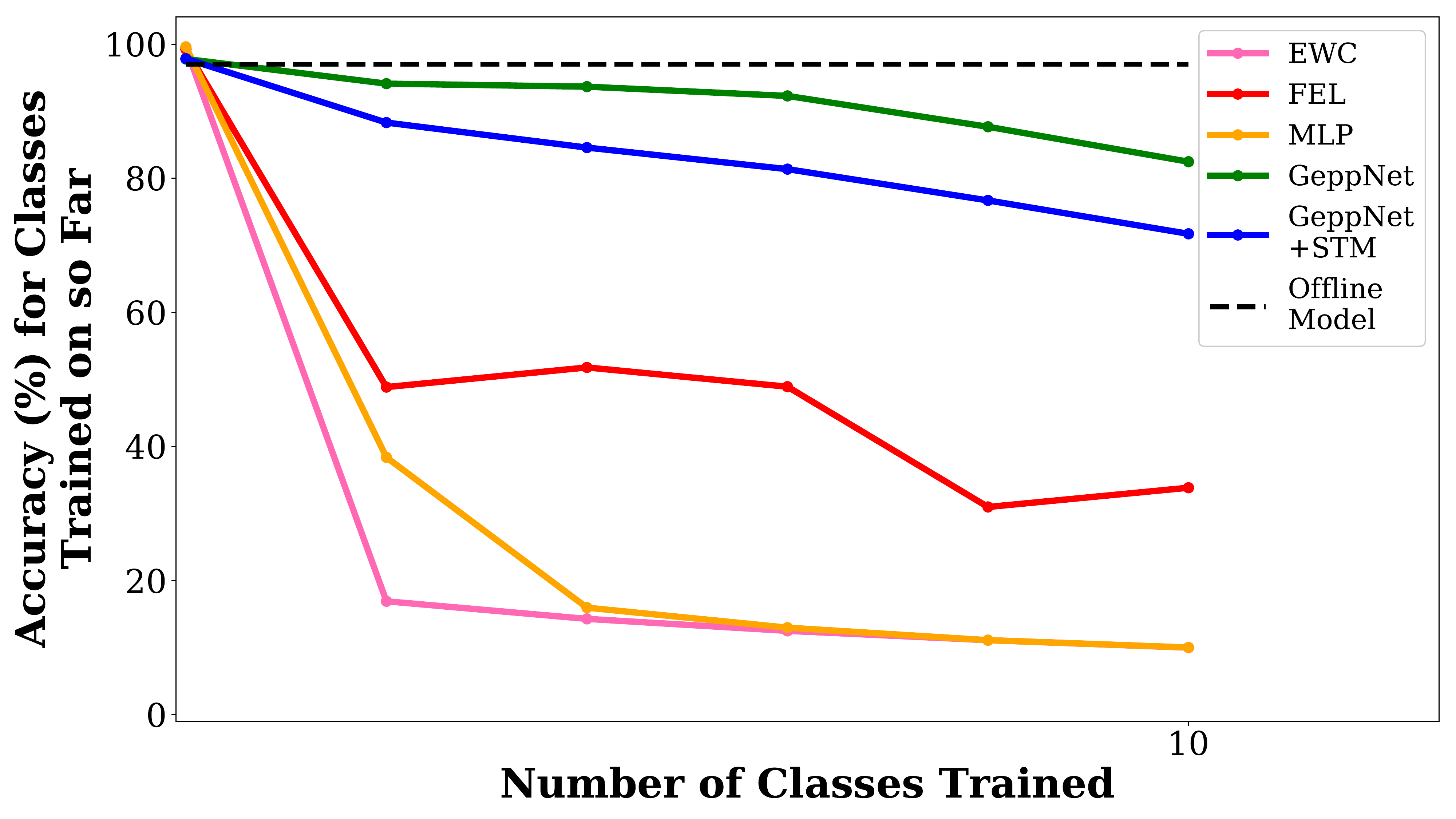}
        \caption{MNIST}
        \label{fig:gull}
    \end{subfigure}

    \begin{subfigure}[b]{0.47\textwidth}
        \includegraphics[width=\textwidth]{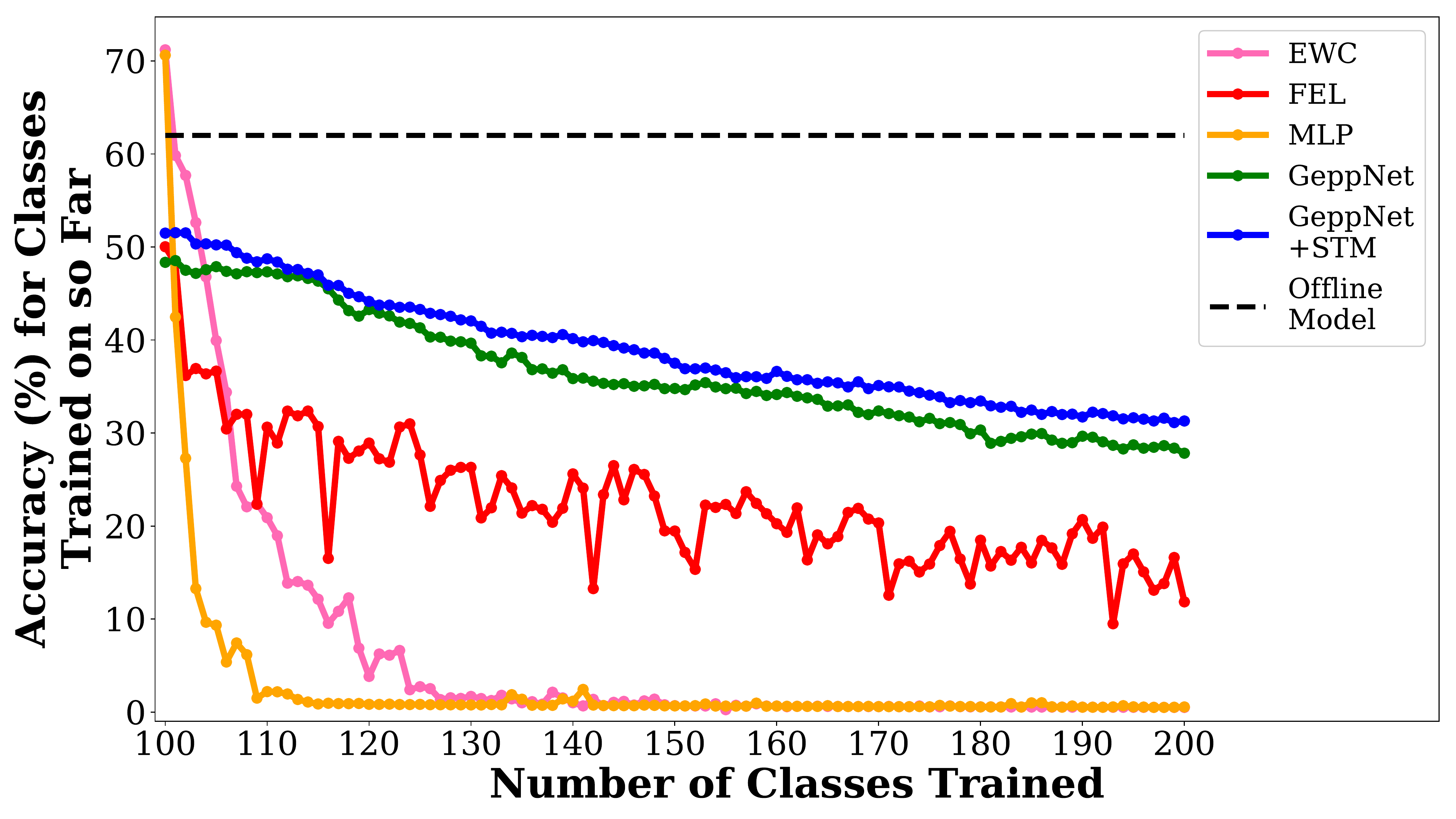}
        \caption{CUB-200}
        \label{fig:tiger}
    \end{subfigure}

    \begin{subfigure}[b]{0.47\textwidth}
        \includegraphics[width=\textwidth]{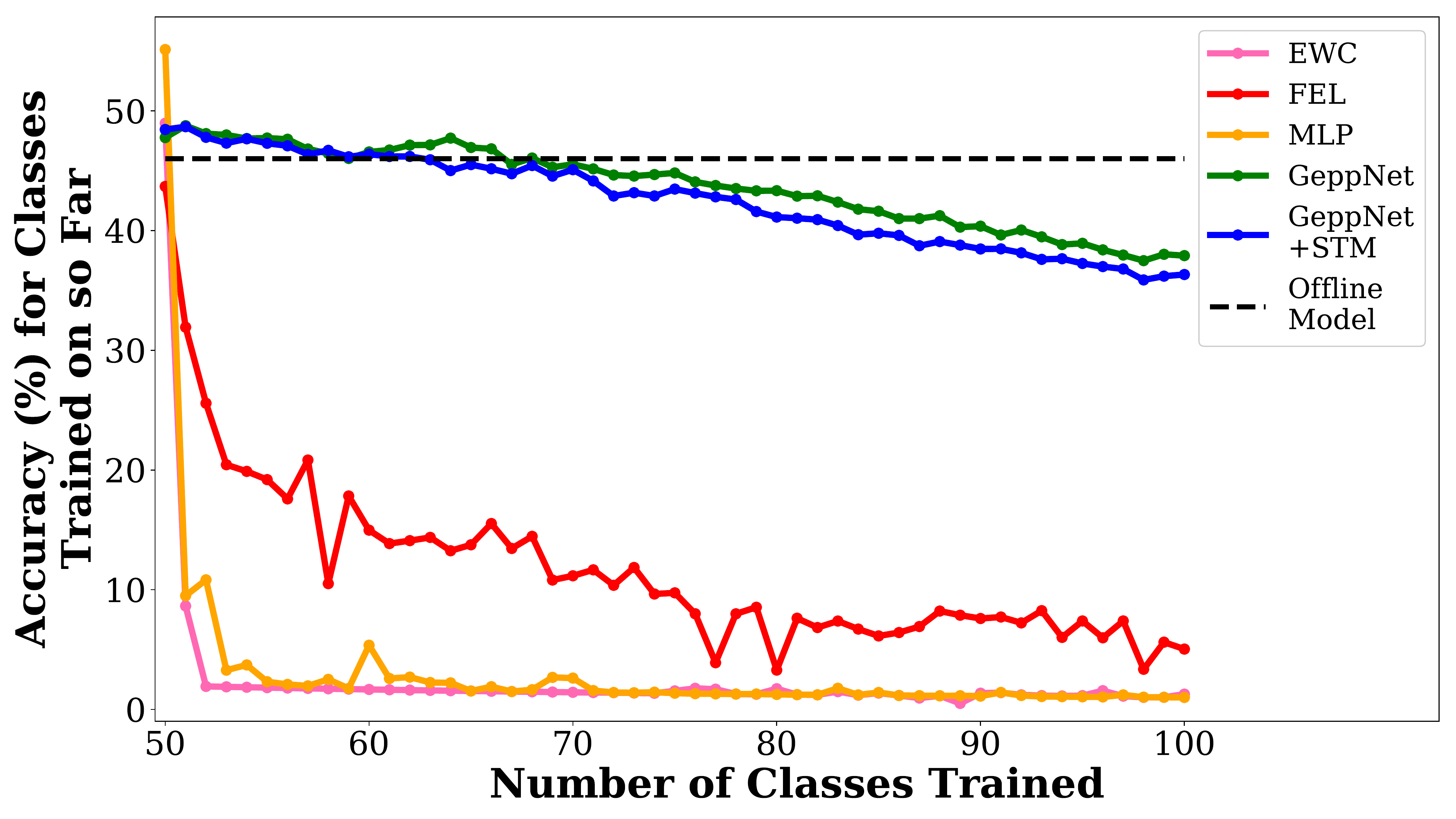}
        \caption{AudioSet}
        \label{fig:mouse}
    \end{subfigure}
    \caption{Results of several lifelong learning approaches for incremental class learning. The mean-class test accuracy evaluated on the MNIST (a), CUB-200 (b), and AudioSet (c) is shown for the following approaches: FEL (red), MLP (yellow), GeppNet (green), GeppNet+STM (blue), EWC (pink), and offline model (dashed line). Adapted with permission from \cite{Kemker2018a}.}\label{fig:tables}
\end{figure*}

In contrast to the datasets typically proposed in the literature to evaluate lifelong learning approaches (e.g., MNIST containing 10 digit classes with low-resolution images; Fig.~\ref{fig:datasets}.a), the above-mentioned experimental conditions were conducted using the Caltech-UCSD Birds-200 (CUB-200) dataset composed of 200 different bird species (\cite{Wah2011}; Fig.~\ref{fig:datasets}.b) and the AudioSet dataset, which is built from YouTube videos with 10-second sound clips from 632 classes and over 2 million annotations~\citep{Gemmeke2017}.
The approaches considered were supervised: a standard MLP trained online as a baseline, the EWC~\citep{Kirkpatrick2017}, the PathNet~\citep{Fernando2017}, the GeppNet and GeppNet+STM~\citep{Gepperth2015}, and the FEL~\citep{Coop2013}.
For the data permutation experiment, best results were obtained by PathNet followed by EWC, suggesting that models that use the ensembling and regularization mechanisms will work best at incrementally learning new tasks/datasets with similar feature distributions.
In contrast, EWC performed better than PathNet on the multi-modal experiment because EWC does a better job on separating non-redundant (i.e., dissimilar) data.
For the incremental learning task, best results were obtained with a combination of rehearsal and dual-memory systems (i.e. GeppNet+STM), yielding gradual adaptation and knowledge consolidation (see Fig.\ref{fig:tables}).
However, since rehearsal requires the storage of raw training examples, pseudo-rehearsal may be a better strategy for future work.

%\begin{figure*}[t]
%\centering
%\includegraphics[width=\textwidth]{LLReview-tables.pdf}
%\caption{Results of several continual learning approaches for the incremental class learning experiment. The mean-class test accuracy evaluated on the MNIST (a), CUB-200 (b), and AudioSet (c) is shown for the following approaches: EWC (pink), FEL (red), MLP (yellow), GeppNet (green), GeppNet+STM (blue), and offline model (dashed line). Adapted from Kemker et al. (2017).}% (\emph{$G^C$}).}
%\label{fig:tables}
%\end{figure*}

\cite{Lomonaco2017} proposed the CORe50, a novel dataset for continuous object recognition that includes 50 classes of objects observed from different perspectives and includes variations in background, illumination, blurring, occlusion, pose, and scale (Fig.~\ref{fig:datasets}.c).
With respect to the above-discussed datasets, CORe50 provides samples collected in experimental conditions closer to what autonomous agents and robots are exposed to in the real world (see Sec. 4).
Along with the dataset, the authors propose three incremental learning scenarios: (i) \textit{new instances} (NI) where all classes are shown in the first batch while subsequent instances of known classes become available over time, \textit{new classes} (NC) where, for each sequential batch, new object classes are available so that the model must deal with the learning of new classes without forgetting previously learned ones, and (iii) \textit{new instances and classes} (NIC) where both new classes and instances are presented in each training batch.
According to the reported results, EWC~\citep{Kirkpatrick2017} and LwF~\citep{Li2016} perform significantly worse in NC and NIC than in NI.

Perhaps not surprisingly, overall performance generally drops when using datasets of higher complexity such as CUB-200 and CORe50 than when tested on the MNIST.
Such results indicate that lifelong learning is a very challenging task and, importantly, that the performance of most approaches can differ significantly according to the specific learning strategy.
This suggests that while there is a large number of approaches capable of alleviating catastrophic forgetting in highly controlled experimental conditions, lifelong learning has not been tackled for more complex scenarios.
Therefore, additional research efforts are required to develop robust and flexible approaches subject to more exhaustive, benchmark evaluation schemes.

\section{Developmental Approaches and Autonomous Agents}

\subsection{Towards Autonomous Agents}

Humans have the extraordinary ability to learn and progressively fine-tune their sensorimotor skills in a lifelong manner~\citep{Tani2016, Bremner2012, Calvert2004}.
Since the moment of birth, humans are immersed in a highly dynamic crossmodal environment which provides a wealth of experiences for shaping perception, cognition, and behaviour~\citep{Murray2016, Lewkowicz2014}.
A crucial component of lifelong learning in infants is their spontaneous capacity of autonomously generating goals and exploring their environment driven by intrinsic motivation~\citep{Cangelosi2015, Gopnik1999}.
Consequently, the ability to learn new tasks and skills autonomously through intrinsically motivated exploration is one of the main factors that differentiate biological lifelong learning from current continual neural networks models of classification.

While there has been significant progress in the development of models addressing incremental learning tasks (see Sec. 3), such models are designed to alleviate catastrophic forgetting from a set of annotated data samples.
Typically, the complexity of the datasets used for the evaluation of lifelong learning tasks is very limited and does not reflect the richness and level of uncertainty of the stimuli that artificial agents can be exposed to in the real world.
Furthermore, neural models are often trained with data samples shown in isolation or presented in a random order.
This significantly differs from the highly organized manner in which humans and animals efficiently learn from samples presented in a meaningful order for the shaping of increasingly complex concepts and skills~\citep{Krueger2009, Skinner1958}.
Therefore, learning in a lifelong manner goes beyond the incremental accumulation of domain-specific knowledge, enabling to transfer generalized knowledge and skills across multiple tasks and domains~\citep{Barnett2002} and, importantly, benefiting from the interplay of multisensory information for the development and specialization of complex neurocognitive functions~\citep{Murray2016, Tani2016, Lewkowicz2014}.

\begin{figure*}[t]
\centering
\includegraphics[width=0.79\textwidth]{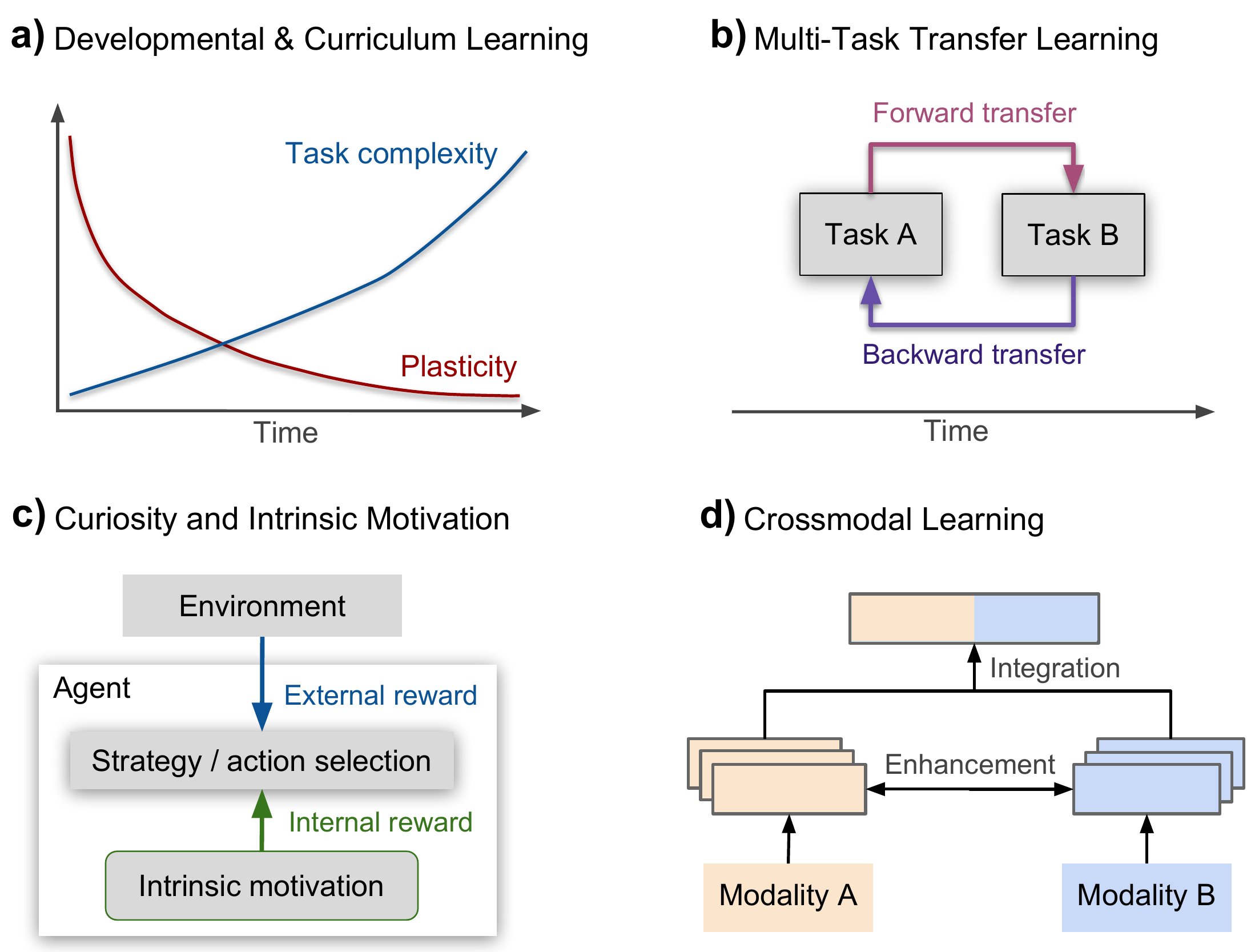}
\caption{Schematic view of the main components for the development of autonomous agents able to learn over long periods of time in complex environments: Developmental and curriculum learning (Sec. 4.2), transfer learning (Sec. 4.3), curiosity and intrinsic motivation (Sec. 4.4), and crossmodal learning (Sec. 4.5).}% (\emph{$G^C$}).}
\label{fig:cca}
\end{figure*}

Intuitively, it is unrealistic to provide an artificial agent with all the necessary prior knowledge to effectively operate in real-world conditions~\citep{Thrun1995}.
Consequently, artificial agents must exhibit a richer set of learning capabilities enabling them to interact in complex environments with the aim to process and make sense of continuous streams of (often uncertain) information~\citep{Hassabis2017, Wermter2005}.
In the last decade, significant advances have been made to embed biological aspects of lifelong learning into neural network models.
In this section, we summarize well-established and emerging neural network approaches driven by interdisciplinary research introducing findings from neuroscience, psychology, and cognitive sciences for the development of lifelong learning autonomous agents.
We focus on discussing models of critical developmental stages and curriculum learning (Sec. 4.2), transfer learning for the reuse of consolidated knowledge during the acquisition of new tasks (Sec. 4.3), autonomous exploration and choice of goals driven by curiosity and intrinsic motivation (Sec. 4.4), and the crossmodal aspects of lifelong learning for multisensory systems and embodied agents (Sec. 4.5).
In particular, we discuss on how these components (see Fig.~\ref{fig:cca}) can be used (independently or combined) to improve current approaches addressing lifelong learning.

\subsection{Developmental and Curriculum Learning}

Learning and development interact in a very intricate way~\citep{Elman1993}.
Humans show an exceptional capacity to learn throughout their lifespan and, with respect to other species, exhibit the lengthiest developmental process for reaching maturity.
There is a limited time window in development in which infants are particularly sensitive to the effects of their experiences.
This period is commonly referred to as the \textit{sensitive} or \textit{critical} period of development~\citep{Lenneberg1967} in which early experiences are particularly influential, sometimes with irreversible effects in behaviour~\citep{Senghas2004}.
During these critical periods, the brain is particularly plastic~(Fig.~\ref{fig:cca}.a) and neural networks acquire their overarching structure driven by sensorimotor experiences (see \cite{Power2016} for a survey).
Afterwards, plasticity becomes less prominent and the system stabilizes, preserving a certain degree of plasticity for its subsequent adaptation and reorganisation at smaller scales~\citep{Hensch1998, Quadrato2014, Kiyota2017}.

The basic mechanisms of critical learning periods have been studied in connectionist models~\citep{Thomas2006, Richardson2008}, in particular with the use of self-organizing learning systems which reduce the levels of functional plasticity through a two-phase training of the topographic neural map~\citep{Kohonen1982, Kohonen1995, Miikkulainen1997}.
In a first organization phase, the neural map is trained with a high learning rate and large spatial neighbourhood size, allowing the network to reach an initial rough topological organization.
In a second tuning phase, the learning rate and the neighbourhood size are iteratively reduced for fine-tuning.
Implementations of this kind have been used to develop models of early visual development~\citep{Miller1989}, language acquisition~\citep{LambonRalph2006, Li2004}, and recovery from brain injuries~\citep{Marchman1993}.
Recent studies on critical periods in deep neural networks showed that the initial rapid learning phase plays a key role in defining the final performance of the networks~\citep{Achille2017}.
The first few epochs of training are critical for the allocation of resources across different layers dictated by the initial input distribution.
After such a critical period, the initially allocated neural resources can be re-distributed through additional learning phases.

Developmental learning strategies have been experimented on with embedded agents to regulate the embodied interaction with the environment in real time~\citep{Cangelosi2015, Tani2016}.
In contrast to computational models that are fed with batches of information, developmental agents acquire an increasingly complex set of skills based on their sensorimotor experiences in an autonomous manner.
Consequently, staged development becomes essential for bootstrapping cognitive skills with less amount of tutoring experience.
However, the use of developmental strategies for artificial learning systems has shown to be a very complex practice.
In particular, it is difficult to select a well-defined set of developmental stages that favours the overall learning performance in highly dynamic environments.
For instance, in the predictive coding framework~\citep{Adams2015, Rao1999}, the intention towards a goal can be generated through the prediction of the consequence of an action by means of the error regression with the prediction error.
The use of generative models, which are implicit in predictive coding, is one component embedded in the framework of active inference~\citep{Friston2015}.
Active inference models aim to understand how to select the data that best discloses its causes in dynamic and uncertain environments through the bilateral use of action and perception.
Nevertheless, it remains unclear how to systematically define developmental stages on the basis of the interaction between innate structure, embodiment, and (active) inference.

Humans and animals exhibit better learning performance when examples are organized in a meaningful way, e.g., by making the learning tasks gradually more difficult~\citep{Krueger2009}.
Following this observation, referred to as \textit{curriculum learning}, \cite{Elman1993} showed that having a curriculum of progressively harder tasks~(Fig.~\ref{fig:cca}.a) leads to faster training performance in neural network systems.
This has inspired similar approaches in robotics~\citep{Sanger1994} and more recent machine learning methods studying the effects of curriculum learning in the performance of learning~\citep{Bengio2009, Reed2015, Graves2016}.
Experiments on datasets of limited complexity (such as MNIST) showed that curriculum learning acts as unsupervised pre-training, leading to improved generalization and faster speed of convergence of the training process towards the global minimum.
However, the effectiveness of curriculum learning is highly sensitive with respect to the modality of progression through the tasks.
Furthermore, this approach assumes that tasks can be ordered by a single axis of difficulty.
\cite{Graves2017} proposed to treat the task selection problem as a stochastic policy over the tasks that maximizes the learning progress, leading to an improved efficiency in curriculum learning.
In this case, it is necessary to introduce additional factors such as intrinsic motivation~\citep{Oudeyer2007, Barto2013}, where indicators of learning progress are used as reward signals to encourage exploration (see Sec. 4.4).
Curriculum strategies can be seen as a special case of transfer learning~\citep{Weiss2016}, where the knowledge collected during the initial tasks is used to guide the learning process of more sophisticated ones.

\subsection{Transfer Learning}

Transfer learning refers to applying previously acquired knowledge in one domain to solve a problem in a novel domain~\citep{Barnett2002, Pan2010, Holyoak1997}.
In this context, forward transfer refers to the influence that learning a task $\mathcal{T}_A$ has on the performance of a future task $\mathcal{T}_B$, whereas backward transfer refers to the influence of a current task $\mathcal{T}_B$ on a previous task $\mathcal{T}_A$~(Fig.~\ref{fig:cca}.b).
For this reason, transfer learning represents a significantly valuable feature of artificial systems for inferring general laws from (a limited amount of) particular samples, assuming the simultaneous availability of multiple learning tasks with the aim to improve the performance at one specific task.

Transfer learning has remained an open challenge in machine learning and autonomous agents (see \cite{Weiss2016} for a survey).
Specific neural mechanisms in the brain mediating the high-level transfer learning are poorly understood, although it has been argued that the transfer of abstract knowledge may be achieved through the use of conceptual representations that encode relational information invariant to individuals, objects, or scene elements~\citep{Doumas2008}.
Zero-shot learning~\citep{Lampert2009, Palatucci2009} and one-shot learning~\citep{FeiFei2003, Vinyals2016} aim at performing well on novel tasks but do not prevent catastrophic forgetting on previously learned tasks.
An early attempt to realize lifelong learning through transfer learning was proposed by \cite{Ring1997} through the use of a hierarchical neural network that solves increasingly complex reinforcement learning tasks by incrementally adding neural units and encode a wider temporal context in which actions take place.

More recent deep learning approaches have attempted to tackle lifelong transfer learning in a variety of domains.
\cite{Rusu2017} proposed the use of progressive neural networks~\citep{Rusu2016} to transfer learned low-level features and high-level policies from a simulated to a real environment.
The task consists of learning pixel-to-action reinforcement learning policies with sparse rewards from raw visual input to a physical robot manipulator.
\cite{Tessler2017} introduced a hierarchical deep reinforcement learning network that uses an array of skills and skill distillation to reuse and transfer knowledge between tasks.
The approach was evaluated by teaching an agent to solve tasks in the Minecraft video game.
However, skill networks need to be pre-trained and cannot be learned along with the overarching architecture in an end-to-end fashion.
\cite{LopezPaz2017} proposed the Gradient Episodic Memory (GEM) model that alleviates catastrophic forgetting and performs \textit{positive} transfer to previously learned tasks.
The model learns the subset of correlations common to a set of distributions or tasks, able to predict target values associated with previous or novel tasks without making use of task descriptors.
However, similar to an issue shared with most of the approaches discussed in Sec. 3, the GEM model was evaluated on the MNIST and CIFAR100 datasets.
Therefore, the question remains whether GEM scales up to more realistic scenarios.

\subsection{Curiosity and Intrinsic Motivation}

Computational models of intrinsic motivation have taken inspiration from the way human infants and children choose their goals and progressively acquire skills to define developmental structures in lifelong learning frameworks~(\cite{Baldassarre2013}; see \cite{Gottlieb2013} for a review).
Infants seem to select experiences that maximize an intrinsic learning reward through an empirical process of exploration~\citep{Gopnik1999}.
From a modelling perspective, it has been proposed that the intrinsically motivated exploration of the environment, e.g., driven by the maximization of the learning progress~(\cite{Oudeyer2007, Schmidhuber1991}, see Fig.~\ref{fig:cca}.c for a schematic view), can lead to the self-organization of human-like developmental structures where the skills being acquired become progressively more complex.

Computational models of intrinsic motivation can collect data and acquire skills incrementally through the online (self-)generation of a learning curriculum~\citep{Baranes2013, Forestier2016}.
This allows the efficient, stochastic selection of tasks to be learned with an active control of the growth of the complexity.
Recent work in reinforcement learning has included mechanisms of curiosity and intrinsic motivation to address scenarios where the rewards are sparse or deceptive~\citep{Forestier2017, Pathak2017, Tanneberg2017, Bellemare2016, Kulkarni2016}.
In a scenario with very sparse extrinsic rewards, curiosity-driven exploration provides intrinsic reward signals that enable the agent to autonomously and progressively learn tasks of increasing complexity.
\cite{Pathak2017} proposed an approach to curiosity-driven exploration where curiosity is modelled as the error in an agent's ability to predict the consequences of its own actions.
This approach has shown to scale up to high-dimensional visual input, using the knowledge acquired from previous experiences for the faster exploration of unseen scenarios.
However, the method relies on interaction episodes that convert unexpected interactions into intrinsic rewards, which does not extend to scenarios where interactions are rare.
In this case, internally generated representations of the previous sparse interactions could be replayed and used to guide exploration (in a similar way to generative systems for memory replay; see Sec. 3.4).

\subsection{Multisensory Learning}

The ability to integrate multisensory information is a crucial feature of the brain that yields a coherent, robust, and efficient interaction with the environment~\citep{Spence2014, Ernst2004, Stein1993}.
Information from different sensor modalities (e.g. vision, audio, proprioception) can be integrated into multisensory representations or be used to enhance unisensory ones (see Fig.~\ref{fig:cca}.d).

Multisensory processing functions are the result of the interplay of the physical properties of the crossmodal stimuli and prior knowledge and expectations (e.g., in terms of learned associations), scaffolding perception, cognition, and behaviour (see \cite{Murray2016, Stein2014} for reviews).
The process of multisensory learning is dynamic across the lifespan and is subject to both short- and long-term changes. 
It consists of the dynamic reweighting of exogenous and endogenous factors that dictate to which extent multiple modalities interact with each other.
Low-level stimulus characteristics (e.g., spatial proximity and temporal coincidence) are available before the formation of learned perceptual representations that bind increasingly complex higher-level characteristics (e.g., semantic congruency).
Sophisticated perceptual mechanisms of multisensory integration emerge during development, starting from basic processing capabilities and progressively specializing towards more complex cognitive functions on the basis of sensorimotor experience~\citep{Lewkowicz2014, Spence2014}.

From a computational perspective, modelling multisensory learning can be useful for a number of reasons.
First, multisensory functions aim at yielding robust responses in the case of uncertain and ambiguous sensory input.
Models of causal inference have been applied to scenarios comprising the exposure to incongruent audio-visual information for solving multisensory conflicts~\citep{Parisi2017b, Parisi2018a}.
Second, if trained with multisensory information, one modality can be reconstructed from available information in another modality.
\cite{Moon2015} proposed multisensory processing for an audio-visual recognition task in which knowledge in a source modality can be transferred to a target modality.
Abstract representations obtained from a network encoding the source modality can be used to fine-tune the network in the target modality, thereby relaxing the imbalance of the available data in the target modality.
\cite{Barros2017} proposed a deep architecture modelling crossmodal expectation learning.
After a training phase with audio-visual information, unisensory network channels can reconstruct the expected output from the other modality.
Finally, mechanisms of attention are essential in lifelong learning scenarios for processing relevant information in complex environments and efficiently triggering goal-directed behaviour from continuous streams of multisensory information~\citep{Spence2014}.
Such mechanisms may be modelled via the combination of the exogenous properties of crossmodal input, learned associations and crossmodal correspondences, and internally generated expectations~\citep{Chen2017} with the aim of continually shaping perception, cognition, and behaviour in autonomous agents.

\section{Conclusion}

Lifelong learning represents an utterly interesting but challenging component of artificial systems and autonomous agents operating on real-world data, which is typically non-stationary and temporally correlated.
%A variety of biological factors of lifelong learning in humans and animals have inspired a wide range of machine learning and neural network approaches that alleviate the problem of catastrophic forgetting.
The mammalian brain remains the best model of lifelong learning, which makes biologically-inspired learning models a compelling approach.
The general notion of structural plasticity (Sec. 2.2) is widely used across the machine learning literature and represents a promising solution to lifelong learning in its own right, even when disregarding biological desiderata.
Proposed computational solutions for mitigating catastrophic forgetting and interference have focused on regulating intrinsic levels of plasticity to protect acquired knowledge~(Sec.~3.2), dynamically allocating new neurons or network layers to accommodate novel knowledge~(Sec.~3.3), and using complementary learning networks with experience replay for memory consolidation~(Sec.~3.4).
However, despite significant advances, current models of lifelong learning are still far from providing the flexibility, robustness, and scalability exhibited by biological systems.
The most popular deep and shallow learning models of lifelong learning are restricted to the supervised domain, relying on large amounts of annotated data collected in controlled environments~(see Sec. 3.5).
Such a domain-specific training scheme cannot be applied directly to autonomous agents that operate in highly dynamic, unstructured environments.

Additional research efforts are required to combine multiple methodologies that integrate a variety of factors observed in human learners.
Basic mechanisms of critical periods of development (Sec. 4.2) can be modelled to empirically determine convenient (multilayered) neural network architectures and initial patterns of connectivity that improve the performance of the model for subsequent learning tasks.
Methods comprising curriculum and transfer learning (Sec. 4.3) are a fundamental feature for reusing previously acquired knowledge and skills to solve a problem in a novel domain by sharing low- and high-level representations.
For agents learning autonomously, approaches using intrinsic motivation (Sec. 4.4) are crucial for the self-generation of goals, leading to an empirical process of exploration and the progressive acquisition of increasingly complex skills.
Finally, multisensory integration (Sec. 4.5) is a key feature of autonomous agents operating in highly dynamic and noisy environment, leading to robust learning and behaviour also in situations of uncertainty.

\section*{Acknowledgment}

This research was partially supported by the German Research Foundation (DFG) under project Transregio Crossmodal Learning (TRR 169).
The authors would like to thank Sascha Griffiths, Vincenzo Lomonaco, Sebastian Risi, and Jun Tani for valuable feedback and suggestions.

\bibliographystyle{agsm}
\bibliography{mybibNN}  % .bib

\end{document}